\documentclass[twocolumn]{svjour3}          
\smartqed  
\usepackage{graphicx}
\usepackage[numbers]{natbib}
\usepackage{times}
\usepackage{epsfig}
\usepackage{amsmath}
\usepackage{amssymb}
\usepackage{lineno}

\usepackage[pagebackref=true,breaklinks=true,letterpaper=true,colorlinks,bookmarks=false,citecolor=blue,linkcolor=blue]{hyperref}

\usepackage{booktabs}
\usepackage{amsmath}
\usepackage{amssymb}
\usepackage{verbatim}
\usepackage{multirow, makecell}
\usepackage{lineno}
\usepackage{enumerate}
\usepackage{amsfonts}
\usepackage{gensymb}
\usepackage{bm}
\usepackage{bbding}
\usepackage{comment}
\usepackage{color}
\usepackage{diagbox}
\usepackage{tabularx}
\usepackage{rotating}
\newcolumntype{L}[1]{>{\raggedright\let\newline\\\arraybackslash\hspace{0pt}}m{#1}}
\newcolumntype{C}[1]{>{\centering\arraybackslash}m{#1}}
\newcolumntype{R}[1]{>{\raggedleft\let\newline\\\arraybackslash\hspace{0pt}}m{#1}}
\newlength\savewidth\newcommand\shline{\noalign{\global\savewidth\arrayrulewidth
  \global\arrayrulewidth 1pt}\hline\noalign{\global\arrayrulewidth\savewidth}}
\newcommand{\tablestyle}[2]{\setlength{\tabcolsep}{#1}\renewcommand{\arraystretch}{#2}\centering\footnotesize}

\usepackage{colortbl}
\definecolor{Gray}{gray}{0.9}
\definecolor{GrayT}{gray}{0.4}
\newcommand{\etal}{{\em et al.}}
\newcommand{\eg}{{\em e.g.}}
\newcommand{\ie}{{\em i.e.}}
\newcommand{\etc}{{\em etc}}
\definecolor{light-blue}{RGB}{186,175,255}
\definecolor{light-red}{RGB}{255,137,165}
\usepackage{arydshln} 
\usepackage{booktabs}
\usepackage{multirow}
\usepackage{amsmath}
\usepackage{etoolbox}
\usepackage{bm}
\usepackage{mathtools}
\usepackage{algorithm}
\usepackage{algpseudocode}
\usepackage{xspace}

\begin{document}
\begin{sloppypar}

\title{HyRSM++: Hybrid Relation Guided Temporal Set Matching for Few-shot Action Recognition
}


\author{Xiang Wang      \and
        Shiwei Zhang    \and
        Zhiwu Qing       \and
        Zhengrong Zuo       \and
        Changxin Gao \and
     Rong Jin \and
     Nong Sang
}


\institute{
Xiang Wang \and Zhiwu Qing \and Zhengrong Zuo \and Changxin Gao (Corresponding author) \and Nong Sang
\at
              Key Laboratory of Ministry of Education for Image Processing and Intelligent Control, School of Artificial
Intelligence and Automation, Huazhong University of Science and Technology  \\
              \email {\{wxiang, qzw, zhrzuo, cgao, nsang\}@hust.edu.cn}           
           \and
           Shiwei Zhang \at
              Alibaba Group \\
              \email {zhangjin.zsw@alibaba-inc.com}
          \and
          Rong Jin \at 
          Twitter \\
          \email {rongjinemail@gmail.com}
}

\date{Received: date / Accepted: date}

\maketitle

\begin{abstract}
Few-shot action recognition is a challenging but practical problem aiming to learn a model that can be easily adapted to identify new action categories with only a few labeled samples.
Recent attempts mainly focus on learning deep representations for each video  individually under the episodic meta-learning regime and then performing temporal alignment to match query and support videos.
However, they still suffer from two drawbacks: (i) learning individual features without considering the entire task may result in limited representation capability, and (ii) existing alignment strategies are sensitive to noises and misaligned instances.
To handle the two limitations, 
we propose a novel Hybrid Relation guided temporal Set Matching (HyRSM++) approach for few-shot action recognition.
The core idea of HyRSM++ is to integrate all videos within the task to learn discriminative representations and involve a robust matching technique.
To be specific, HyRSM++ consists of two key components, a hybrid relation module and a temporal set matching metric.
Given the basic representations from the feature extractor, the hybrid relation module is introduced to fully exploit associated relations within and cross videos in an episodic task and thus can learn task-specific embeddings.
Subsequently, in the temporal set matching metric, we carry out the distance measure between query and support videos from a set matching perspective and design a bidirectional Mean Hausdorff Metric to improve the resilience to misaligned instances. 
In addition, we explicitly exploit the temporal coherence in videos to regularize the matching process.
In this way, HyRSM++ facilitates informative correlation exchanged among videos and enables flexible predictions under the data-limited scenario.
Furthermore, we extend the proposed HyRSM++ to deal with the more challenging semi-supervised few-shot action recognition and unsupervised few-shot action recognition tasks. 
%
Experimental results on multiple benchmarks demonstrate that our method consistently outperforms existing methods and achieves state-of-the-art performance under various few-shot settings. The source code is available at \url{https://github.com/alibaba-mmai-research/HyRSMPlusPlus}.


\end{abstract}

\section{Introduction}
\label{intro}

{Recently}, the development of large-scale video benchmarks~\citep{Kinetics,SSV2,caba2015activitynet,EPIC-100,grauman2022ego4d} and deep networks~\citep{TSN,TSM,Slowfast,SSTAP,peng2018unsupervised,liu2015learning} have significantly boosted the progress of action recognition.
To achieve this success, we typically require large amounts of manually labeled data.
However, acquiring these labeled examples consumes a lot of manpower and time, which actually limits further applications of this task.
In this case, researchers look to alternatives to achieve action classification without extensive costly labeling.
%
Few-shot action recognition is a promising direction to reduce manual annotations and thus has attracted much attention recently~\citep{CMN,ARN-ECCV}.
It aims at learning to classify unseen action classes with extremely few annotated examples. 

\begin{figure*}[t]  
\centering
\includegraphics[width=0.95\textwidth]{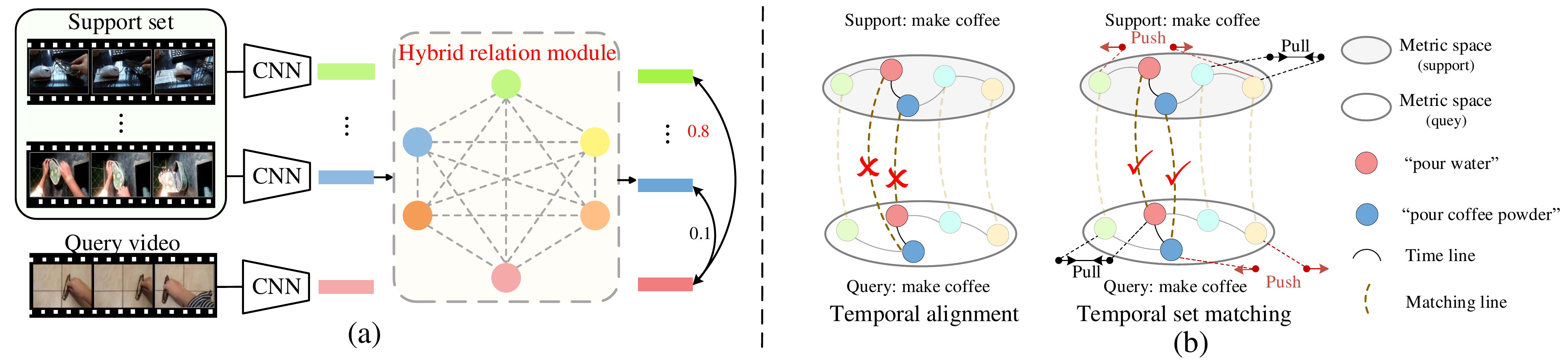}
\vspace{-1mm}
\caption{
%
%
(a) Concept of the proposed hybrid relation module.
We adaptively produce  task-specific  video embeddings by extracting relevant discriminative patterns cross videos in an episodic task.
(b) Example of \emph{make coffee}, the current temporal alignment metrics tend to be strict, resulting in an incorrect match on misaligned videos.
In contrast, the proposed temporal set matching metric involving set matching technique and temporal coherence regularization is more flexible in finding the best correspondences.
}
\label{fig:Motivation}
\end{figure*}
%
%
%
%
%
To solve the few-shot data-scarcity problem, popular attempts~\citep{CMN,OTAM,TRX,ITANet} are mainly based on the metric-based meta-learning technique~\citep{MatchNet}, in which a common embedding space is first learnt via episodic training and then an explicit or implicit alignment metric is employed to calculate the distances between the query (test) videos and support (reference) videos for classification in an episodic task.
Typically, Ordered Temporal Alignment Module (OTAM)~\citep{OTAM} adopts a deep feature extractor to convert an input video into a frame feature sequence independently and explicitly explores the ordered temporal alignment path between support and query videos in this feature space.
Temporal-Relational CrossTransformer (TRX)~\citep{TRX} learns a deep embedding space and tries to exhaustively construct temporally-corresponding sub-sequences of actions to compare.
Some recent works~\citep{huang2022compound,MTFAN,zheng2022few,nguyen2022inductive} propose to design multi-level metrics for few-shot action recognition.

Although these methods have achieved remarkable performance, there are still two limitations: individual feature learning and inflexible matching strategy.
First, discriminative interactive clues cross videos in an episode are ignored when each video is considered independently during representation learning.
As a result, these methods actually assume the learned representations are equally effective on different episodic tasks and maintain a fixed set of video features for all test-time tasks, \ie, task-agnostic, which hence might overlook the most discriminative dimensions for the current task.
Existing work also shows that the task-agnostic methods tend to suffer inferior generalization in other fields, such as image recognition~\citep{Finding_task-relevant, TapNet}, NLP~\citep{NLP_task-specific-1, NLP_task-specific-2}, and information retrieval~\citep{Retrieval_task-specific-1}.
Second, actions are usually complicated and involve many subactions with different orders and offsets, which may cause the failure of existing temporal alignment metrics.
For example, as shown in Figure~\ref{fig:Motivation}(b), to \emph{make coffee}, you can \emph{pour water} before \emph{pour coffee powder}, or in a reverse order, hence it is hard for recent temporal alignment strategies to find the right correspondences.
Thus a more flexible metric is required to cope with the misalignment.
%

%
Inspired by the above observations, we thus solve the few-shot action recognition problem by developing a novel Hybrid Relation guided temporal Set Matching algorithm, dubbed HyRSM++, which is architecturally composed of a hybrid relation module and a temporal set matching metric.
In the hybrid relation module, we argue that the considerable relevant relations within and cross videos are beneficial to generate a set of customized features that are discriminative for a given task.
%
To this end, we first apply an intra-relation function to strengthen structural patterns  within a video via modeling long-range temporal dependencies.
Then an inter-relation function operates on different videos to extract rich semantic information to reinforce the features which are more relevant to query predictions, as shown in Figure~\ref{fig:Motivation}(a).
By this means, we can learn task-specific embeddings for the few-shot task.
%
%
%
On top of the hybrid relation module, we design a novel temporal set matching metric consisting of a bidirectional Mean Hausdorff Metric and a temporal coherence regularization to calculate the distances between query and support videos, as shown in Figure~\ref{fig:Motivation}(b).
The objective of the bidirectional Mean Hausdorff Metric is to measure video distance from the set matching perspective.
Concretely, we treat each video as a set of frames and alleviate the strictly ordered constraints to acquire better query-support correspondences.
Furthermore, to exploit long-range temporal order dependencies, we explicitly impose temporal coherence regularization on the input videos for more stable measurement without introducing extra network parameters.
In this way,  by combining the hybrid relation module and temporal set matching metric, the proposed HyRSM++ can sufficiently integrate semantically relational representations within the entire task and provide flexible video matching in an end-to-end manner.
We evaluate the proposed HyRSM++ on six challenging benchmarks and achieve remarkable improvements again current state-of-the-art methods.
%

%
Although the intuition of HyRSM++ is straightforward, it is elaborately designed for few-shot action recognition. 
Can our HyRSM++ be applied to the more challenging semi-supervised or unsupervised action recognition tasks even if the settings are entirely different? 
To answer this question, we extend HyRSM++ to the semi-supervised and unsupervised objectives with minor task adaptation modifications, and experimental results indicate that HyRSM++ can be well adapted to different scenarios well and achieves impressive performance.
%
%

In summary, we make the following four contributions:

 (1) We propose a novel hybrid relation module to capture the intra- and inter-relations inside the episodic task, yielding task-specific representations for different tasks.
 
 (2) We reformulate the query-support video pair distance metric as a set matching problem and develop a bidirectional Mean Hausdorff Metric, which can be robust to complex actions. To utilize long-term temporal order cues, we further design a new temporal coherence  regularization on videos without adding network parameters.
 
 (3) We conduct extensive experiments on six challenging datasets to verify that the proposed HyRSM++ achieves superior performance over the state-of-the-art methods.
 
 (4) We show that the proposed HyRSM++ can be directly extended to the more challenging semi-supervised few-shot action recognition and unsupervised few-shot action recognition task with minor modifications.

In this paper, we have extended our preliminary CVPR-2022 conference version~\citep{HyRSM} in the following aspects.
i) We integrate the temporal coherence  regularization and set matching strategy into a temporal set matching metric so that the proposed metric can explicitly leverage temporal order information in videos and match flexibly.
Note that temporal coherence regularization does not introduce additional parameters and will not increase the burden of inference.
ii) We conduct more comprehensive ablation studies to verify the effectiveness and efficiency of the proposed HyRSM++.
iii) We clearly improve the few-shot action recognition performance over the previous version. Experimental results also manifest that HyRSM++ significantly surpasses existing competitive methods and achieves state-of-the-art performance.
iv) We show that the proposed HyRSM++ can be easily extended to the more challenging semi-supervised few-shot recognition and unsupervised few-shot action recognition tasks.

\section{Related Work}
\label{sec:related}
In the literature, there are some techniques related to this paper, mainly including few-shot image classification, set matching, temporal coherence, semi-supervised few-shot learning, unsupervised few-shot learning, and few-shot action recognition. In this section, we will briefly review them separately. 

\vspace{+1pt}
\noindent \textbf{Few-shot Image Classification. } 
Recently, the research of few-shot learning~\citep{few-shot_feifei,lu2020learning,lu2020robust} has proceeded roughly along with the following directions: data augmentation, optimization-based, and metric-based.
Data augmentation is an intuitive method to increase the number of training samples and improve the diversity of data. 
Mainstream strategies include spatial deformation~\citep{augmentation_1,augmentation_2} and semantic feature augmentation~\citep{augmentation_feature_1,augmentation_feature_2}. 
Optimization-based methods learn a meta-learner model that can quickly adopt to a new task given a few training examples.
These algorithms include the LSTM-based meta-learner~\citep{meta-learner_1}, learning efficient model initialization~\citep{MAML}, and learning stochastic gradient descent optimizer~\citep{meta-learner_3}.
Metric-based methods attempt to address the few-shot classification problem by "learning to compare". 
This family of approaches aims to learn a feature space and compare query and support images through Euclidean distance~\citep{prototypical,TapNet,ye2022few}, cosine similarity~\citep{MatchNet,ye2020few}, or learnable non-linear metric~\citep{RelationNet,Cross_attention,Finding_task-relevant}.
Our work is more closely related to the metric-based methods~\citep{Finding_task-relevant,TapNet} that share the same spirit of learning task-specific features, whereas we focus on solving the more challenging few-shot action recognition task with diverse spatio-temporal dependencies.
%
In addition, we will further point out the differences and conduct performance comparisons in the experimental section.
%

\vspace{+1pt}
\noindent \textbf{Set Matching. }
The objective of set matching is to accurately measure the similarity of two sets, which have received much attention over the years.
Set matching techniques can be used to efficiently process complex data structures~\citep{setmatching_graph,setmatching_cluster,setmatching_graph_2} and has been applied in many computer vision fields, including face recognition~\citep{setmatching_face_1,setmatching_face_2,setmatching_face_3}, object matching~\citep{setmatching_object,Hausdorff_image_4}, \etc.
Among them, Hausdorff distance is an important alternative to handle set matching problems.
Hausdorff distance and its variants have been widely used in the field of image matching and achieved remarkable results~\citep{Hausdorff_image_1,Hausdorff_image_2,Hausdorff_image_3,Hausdorff_image_4,Hausdorff_image_5,Hausdorff_image_6}.
Inspired by these great successes, we introduce set matching into the few-shot action recognition field for the first time.
\vspace{+1pt}
\noindent \textbf{Temporal Coherence. } 
Videos naturally involve temporal continuity, and there is much effort to effectively explore how to leverage this property~\citep{IDM,coherence_unsupervised_1,coherence_unsupervised_2,mitra2016bayesian}.
Inverse Difference Moment (IDM)~\citep{IDM} is a commonly used measure of local homogeneity, which assumes that in a sequence, two elements are more similar if they are located next to each other.
The idea of IDM has been widely applied to texture feature extraction~\citep{mohanaiah2013image}, face recognition~\citep{mobahi2009deep}, and unsupervised representation learning~\citep{coherence_unsupervised_1,coherence_unsupervised_2} and achieved remarkable performance.
In this paper, we focus on constraining the few-shot matching process by exploiting temporal coherence.

\vspace{+1pt}
\noindent \textbf{Semi-supervised Few-shot Learning. } 
In practical application scenarios, there are usually many unlabeled samples.
Semi-supervised few-shot learning considers learning new concepts in the presence of extra unlabeled data.
Ren \etal~\citep{semi-few-shot-1} first introduce the challenging semi-supervised few-shot learning paradigm and refine the prototypes by adopting a soft k-means on unlabeled data.
LST~\citep{semi-few-shot-2} proposes a novel recursive-learning-based self-training strategy for robust convergence of the inner loop.
TransMatch \citep{Transmatch} develops a new transfer learning framework by incorporating MixMatch~\citep{Mixmatch} and existing few-shot learning methods.
PTN~\citep{PTN} employs the Poisson learning model to obtain informative presentations between the labeled and unlabeled data.
PLCM~\citep{PLCM} and iLPC~\citep{iLPC} focus on cleaning predicted pseudo-labels and generating accurate confidence estimation.
In the field of semi-supervised few-shot action recognition, LIM~\citep{CMN-J} utilizes a label-independent memory to preserve a feature bank and produces class prototypes for query classification. 

\vspace{+1pt}
\noindent \textbf{Unsupervised Few-shot Learning. } 
The objective of unsupervised few-shot learning is to utilize unlabeled samples to construct meta-tasks for few-shot training.
CACTUs~\citep{CACTUs} and UFLST~\citep{ji2019unsupervised} construct many tasks by clustering embeddings and optimize the meta-learning process over the constructed tasks.
UMTRA~\citep{UMTRA} generates artificial tasks by randomly sampling support examples from the training set and produces corresponding queries by augmentation.
ULDA~\citep{ULDA} and AAL~\citep{AAL} follow this paradigm to randomly group augmented images for meta-learning and point out the importance of data augmentation.
More recently, MetaUVFS~\citep{MetaUVFS} presents the first unsupervised meta-learning algorithm for few-shot action recognition and adopts a two-stream 2D and 3D CNN model to explore spatial and temporal features via contrastive learning.

%
%
%
%
%
\begin{figure*}[t]  
\centering
\includegraphics[width=0.97\textwidth]{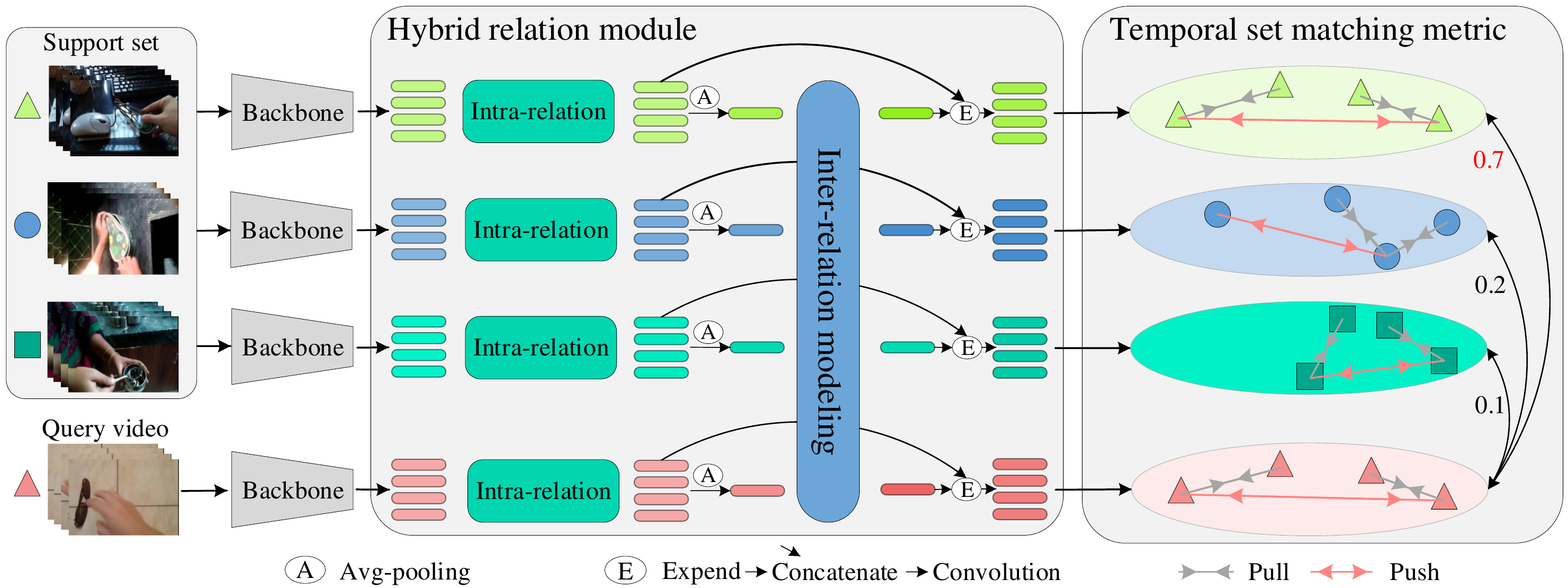}
\caption{Schematic illustration of the proposed Hybrid Relation guided temporal Set Matching (HyRSM++) approach on a 3-way 1-shot problem. 
Given an episode of video data, a feature embedding network is first employed to extract their feature vectors.
Then, A hybrid relation module is followed to integrate rich information within each video and cross videos with intra-relation and inter-relation functions.
%
%
Finally, the task-specific features are fed forward into a temporal set matching metric for matching score prediction.
%
Best viewed in color.
}
\label{fig:Network}
\end{figure*}
%
%
%
%
%

\vspace{+1pt}
\noindent \textbf{Few-shot Action Recognition. }
The difference between few-shot action recognition and the previous few-shot learning approaches is that it deals with more complex higher dimensional video data instead of two-dimensional images.
The existing methods mainly focus on metric-based learning.
OSS-Metric Learning~\citep{OSS-metric} adopts OSS-Metric of video pairs to match videos.
%
TARN~\citep{TARN} learns an attention-based deep-distance measure from an attribute to a class center for zero-shot and few-shot action recognition.
%
CMN~\citep{CMN} utilizes a multi-saliency embedding algorithm to encode video representations.
AMeFu-Net~\citep{few-shot-depth} uses depth information to assist learning.
Xian \etal~\citep{xian2021generalized} propose to learn a generative adversarial network
and produce video features of novel classes for generalization.
Coskun~\etal~\citep{coskun2021domain} leverage object-object interaction, hand grasp, optical flow, and
hand trajectory to learn an egocentric few-shot classifier.
OTAM~\citep{OTAM} preserves the frame ordering in video data and estimates distances with ordered temporal alignment.
ARN~\citep{ARN-ECCV} introduces a self-supervised permutation invariant strategy for spatio-temporal modeling. 
ITANet~\citep{ITANet} proposes a frame-wise implicit temporal alignment strategy to achieve accurate and robust video matching.
TRX~\citep{TRX} matches actions by matching plentiful tuples of different sub-sequences.
More recently, STRM~\citep{STRM} makes use of local and global enrichment mechanism for spatio-temporal modeling based on TRX~\citep{TRX} and enforces class-separability at different phase. 
Some works~\citep{huang2022compound,MTFAN,zheng2022few,nguyen2022inductive} propose to design multi-level metrics for few-shot action recognition.
Note that most above methods focus on learning video embedding independently.
%
Unlike these previous methods, our HyRSM++ improves the transferability of embedding by learning intra- and inter-relational patterns that can better generalize to unseen classes.

\section{Method} 
\label{sec:method}
In this section, we first formulate the definition of the few-shot action recognition task.
Then we present our Hybrid  Relation  guided  temporal Set  Matching  (HyRSM++) method.

\subsection{Problem formulation} 
\label{subsec:definition}
Few-shot action recognition aims to obtain a model that can generalize well to new classes when limited labeled video data is available.
To make training more faithful to the test environment, we adopt the episodic training manner~\citep{MatchNet} for few-shot adaptation as in previous work~\citep{MatchNet,OTAM,TRX,ITANet}.
In each episodic task, there are two sets, \emph{i.e.}, a support set $S$ and a query set $Q$.
The support set $S$ contains $N \times K$ samples from $N$ different action classes, and each class contains $K$ support videos, termed the $N$-way $K$-shot problem.
The goal is to classify the query videos in $Q$ into $N$ classes with these support videos.

\subsection{HyRSM++}
\label{HyRSM++}

\textbf{Pipeline.}
The overall architecture of HyRSM++ is illustrated in Figure~\ref{fig:Network}. 
For each input video sequence, we first divide it into $T$ segments and extract a snippet from each segment, as in previous methods~\citep{TSN,OTAM}.
This way, in an episodic task, the support set can be denoted as $S=\{s_{1}, s_{2}, ..., s_{N \times K}\}$, where $s_{i} = \{s_{i}^{1}, s_{i}^{2}, ...,s_{i}^{T}\}$.
For simplicity and convenience, we discuss the process of the $N$-way $1$-shot problem, \ie, $K=1$, and consider that the query set $Q$ contains a single video $q$.
Then we apply an embedding model to extract the feature representations for each video sequence and obtain the support features $F_s=\{f_{s_1}, f_{s_2},...,f_{s_{N}}\}$ and the query feature $f_q$, where $f_{s_i}=\{f_{i}^{1}, f_{i}^{2}, ...,f_{i}^{T}\}$ and $f_q=\{f_{q}^{1}, f_{q}^{2}, ...,f_{q}^{T}\}$.
After that, we input $F_s$ and $f_q$ to the hybrid relation module to learn task-specific features, resulting in $\tilde{F}_s$ and $\tilde{f}_q$.
Finally, the enhanced representations  $\tilde{F}_s$ and $\tilde{f}_q$ are fed into the set matching metric to generate matching scores.
Based on the output scores, we can train or test the total framework.

\textbf{Hybrid relation module. }
Given the features $F_s$ and $f_q$ output by the embedding network, current approaches, \eg, OTAM~\citep{OTAM}, directly apply a classifier $\mathcal{C}$ in this feature space.
They can be formulated as:
\begin{equation}
y_{i} = \mathcal{C}(f_{s_i}, f_q)
\label{eq:prb_baseline}
\end{equation}
where $y_i$ is the matching score between $f_{s_{i}}$ and $f_q$.
During training, $y_i=1$ if they belong to the same class, otherwise $y_i=0$.
In the testing phase, $y_i$ can be adopted to predict the query label.
From the perspective of probability theory, it makes decisions based on the priors $f_{s_i}$ and $f_q$:
%
\begin{equation}
y_i=\mathcal{P}((f_{s_i}, f_q)|f_{s_i}, f_q)
\label{eq:prb_baseline_pro}
\end{equation}
which is a typical task-agnostic method.
%
%
%
However, the task-agnostic embedding is often vulnerable to overfit irrelevant representations~\citep{Cross_attention, Finding_task-relevant} and may fail to transfer to unseen classes not yet observed in the training stage.
%

%
%
Unlike the previous methods, we propose to learn task-specific features for each target task.
To achieve this goal, we introduce a hybrid relation module to generate task-specific features by capturing rich information from different videos in an episode.
Specifically,  we elaborately design the hybrid relation module $\mathcal{H}$ in the following form:
\begin{equation}
\tilde{f}_i=\mathcal{H}(f_i, \mathcal{G}); f_i \in  [F_s, f_q], \mathcal{G} = [F_s, f_q]
\end{equation}
That is, we improve the feature $f_i$ by aggregating semantic information cross video representations, \ie, $\mathcal{G}$, in an episodic task, allowing the obtained task-specific feature $\tilde{f}_i$ to be more discriminative than the isolated feature. 
For efficiency, we further decompose hybrid relation module into two parts:
intra-relation function $\mathcal{H}_a$ and inter-relation function $\mathcal{H}_e$.
%
%

%
The intra-relation function aims to strengthen structural patterns within a video by capturing long-range temporal dependencies.
We express this process as:
\begin{equation}
    f_i^a = \mathcal{H}_{a}(f_i)
\end{equation}
here $f_i^a \in \mathcal{R}^{T\times C}$ is the output of $f_i$ through the intra-relation function and has the same shape as $f_i$.
Note that the intra-relation function has many alternative implements, including multi-head self-attention (MSA), Transformer~\citep{Transformer}, Bi-LSTM~\citep{BILSTM}, Bi-GRU~\citep{GRU}, \etc., which is incredibly flexible and can be any one of them.

Based on the features generated by the intra-relation function, an inter-relation function is deployed to semantically enhance the features cross different videos:
%
%
\begin{equation}
f_i^e = \mathcal{H}_i^e(f_i^a, \mathcal{G}^a) \ 
        = \sum_j^{|\mathcal{G}^a|}(\kappa(\psi(f_i^a), \psi(f_j^a)) * \psi(f_j^a))
\end{equation}
where $\mathcal{G}^a = [F_s^a, f_q^a]$, $\psi(\cdot)$ is a global average pooling layer, and $\kappa(f_i^a, f_j^a)$ is a learnable function that calculates the semantic correlation between $f_i^a$ and $f_j^a$.
The potential logic is that if the correlation score between $f_i^a$ and $f_j^a$ is high, \ie, $\kappa(f_i^a, f_j^a)$, it means they tend to have the same semantic content, hence we can borrow more information from $f_j^a$ to elevate the representation $f_i^a$, and vice versa.
In the same way, if the score $\kappa(f_i^a, f_i^a)$ is less than $1$, it indicates that some irrelevant information in $f_i^a$ should be suppressed.
In this way, we can improve the feature discrimination by taking full advantage of the limited samples in each episodic task. 
%
%
The inter-relation function also has similar implements with the intra-relation function but with a different target.
After the inter-relation function, we employ an Expend-Concatenate-Convolution operation to aggregate information, as shown in Figure~\ref{fig:Network}, where the output feature $\tilde{f_{i}}$ has the same shape as $f_i^e$.
In the form of prior, our method can be formulated as:
\begin{equation}
y_i=\mathcal{P}((\tilde{f}_{s_i}, \tilde{f_q})|\mathcal{H}(f_{s_i}, \mathcal{G}), \mathcal{H}(f_q, \mathcal{G})); \mathcal{G} = [F_s, f_q]
\label{eq:task_speci_prob}
\end{equation}
Intuitively, compared with Equation~\ref{eq:prb_baseline_pro}, it can be conducive to making better decisions because more priors are provided.
In particular, the hybrid relation module is a plug-and-play unit.
In the experiment, we will fully explore different configurations of the hybrid relation module and further investigate its insertablility.

\textbf{Temporal set matching metric. }
%
%
Many prior few-shot action recognition algorithms usually impose a strict temporal alignment strategy on generated video representations for few-shot classification.
However, they suffer from causing some failed matches when encountering misaligned video instances.
Instead, we develop a flexible metric based on set matching that explicitly discovers optimal frame matching pairs for the ability to be insensitive to misalignment.
Concretely, the proposed temporal set matching metric contains two parts, bidirectional Mean Hausdorff Metric (Bi-MHM) and temporal coherence regularization, respectively. We will describe them in detail below.
%
%

%
%
Given the relation-enhanced features $\tilde{F_{s}}$ and $\tilde{f_{q}}$, we present a novel metric to enable efficient and flexible matching.
In this metric, we treat each video as a set of $T$ frames and reformulate distance measurement between videos as a set matching problem, which is robust to complicated instances, whether they are aligned or not.
Specifically, we achieve this goal by modifying the Hausdorff distance, which is a typical set matching approach.
The standard Hausdorff distance $\mathcal{D}$ can be formulated as:
\begin{equation}
  \begin{split}
   d(\tilde{f_i},\tilde{f_q}) &= \max_{\tilde{f_i^a}\in \tilde{f_i}}(\min_{\tilde{f_q^b}\in \tilde{f_q}} \begin{Vmatrix} \tilde{f_i^a}-\tilde{f_q^b} \end{Vmatrix}) \\
   d(\tilde{f_q},\tilde{f_i}) &= \max_{\tilde{f_q^b}\in \tilde{f_q}}(\min_{\tilde{f_i^a}\in \tilde{f_i}} \begin{Vmatrix} \tilde{f_q^b}-\tilde{f_i^a} \end{Vmatrix}) \\
   \mathcal{D} &= \max(d(\tilde{f_i},\tilde{f_q}), d(\tilde{f_q},\tilde{f_i}))
  \end{split}
\end{equation}
where $\tilde{f_i} \in \mathcal{R}^{T \times C}$ contains $T$ frame features, and $\begin{Vmatrix} \cdot \end{Vmatrix}$ is a distance measurement function, which is the cosine distance in our method.
%

%
However, the previous methods~\citep{hausdorff-modified,hausdorff-modified-1,hausdorff-modified-2,Hausdorff_image_2} pointed out that Hausdorff distance can be easily affected by noisy examples, resulting in inaccurate measurements. 
Hence they employ a directed modified Hausdorff distance that robust to noise as follows:
\begin{equation}
    d_{m}(\tilde{f_i},\tilde{f_q}) = \frac{1}{N_i} \sum_{\tilde{f_i^a} \in \tilde{f_i}}(\min_{\tilde{f_q^b} \in \tilde{f_q}} 
    \begin{Vmatrix}
       \tilde{f_i^a}-\tilde{f_q^b}
   \end{Vmatrix}) 
\end{equation}
where $N_i$ is the length of $\tilde{f_i}$, and equal to $T$ in this paper.
%
Hausdorff distance and its variants achieve great success in image matching~\citep{Hausdorff_image_5,Hausdorff_image_2,Hausdorff_image_1} and face recognition~\citep{hausdorff-modified-1, Hausdorff_image_6}. We thus propose to introduce the set matching strategy into the few-shot action recognition field and further design a novel bidirectional Mean Hausdorff Metric (Bi-MHM):
%
\begin{equation}
\label{equ:bimhd}
 \begin{split}
    \mathcal{D}_b = \frac{1}{N_i} \sum_{\tilde{f_i^a} \in \tilde{f_i}}(\min_{\tilde{f_q^b}\in \tilde{f_q}} \begin{Vmatrix}
       \tilde{f_i^a}-\tilde{f_q^b}
   \end{Vmatrix})  
   + \\ \frac{1}{N_q} \sum_{\tilde{f_q^b} \in \tilde{f_q}}(\min_{\tilde{f_i^a} \in \tilde{f_i}} \begin{Vmatrix}
       \tilde{f_q^b} - \tilde{f_i^a}
   \end{Vmatrix})
 \end{split}
\end{equation}
where $N_i$ and $N_q$ are the lengths of the support feature $\tilde{f_i}$ and the query feature $\tilde{f_q}$ respectively.
%

The proposed Bi-MHM is a symmetric function, and the two items are complementary to each other.
From Equation~\ref{equ:bimhd}, we can find that $\mathcal{D}_b$ can automatically find the best correspondencies between two videos, \eg, $\tilde{f_i}$ and $\tilde{f_q}$.
Note that our Bi-MHM is a non-parametric classifier and does not involve numerous non-parallel calculations, which helps to improve computing efficiency and transfer ability compared to the previous complex alignment classifiers~\citep{OTAM,TRX}.
Moreover, the hybrid relation module and Bi-MHM can mutually reinforce each other, consolidating the correlation between two videos collectively. 

%
%
The Bi-MHM approach described above assumes video sequence representations belonging to the same action have the same set structure in the feature space and does not explicitly utilize temporal order information. 
However, it would be much more general to take the inherent temporal information in videos into account.
%
%
For this reason, we take advantage of the temporal coherence that naturally exists in sequential video data and construct a temporal coherence regularization to further constrain the matching process by incorporating temporal order information.

%
IDM~\citep{IDM} is a commonly used means that can exploit temporal coherence within videos, which can be formulated as:
\begin{equation}
\label{IDM}
I(\tilde{f_i}) = \sum_{a=1}^{T}\sum_{b=1}^{T}  \frac{1}{(a-b)^2+1}\cdot\begin{Vmatrix}\tilde{f_i^a}- \tilde{f_i^b} \end{Vmatrix}
\end{equation}
where $\tilde{f_i}$ is the input video feature, $T$ is the temporal length of the video, and the above loss encourages frames that are close in time to be close in the feature space as well.
In addition, there is another way to use temporal order information in the literature~\citep{coherence_unsupervised_1,mobahi2009deep}:
\begin{equation}
\label{IDM_2}
I(\tilde{f_i};\tilde{f_i^a},\tilde{f_i^b}) =
\begin{cases}
  \begin{Vmatrix}\tilde{f_i^a}- \tilde{f_i^b} \end{Vmatrix}, & \text{ if } \left | a-b \right |=1  \\
 max(0, m-\begin{Vmatrix}\tilde{f_i^a}- \tilde{f_i^b} \end{Vmatrix}) & \text{ if } \left | a-b \right |>1
\end{cases}
\end{equation}
where $m$ is the size of the margin. Equation~\ref{IDM_2} utilizes the video coherence property by pulling two frame features closer if they are adjacent, pushing farther apart by one margin $m$ if they are not adjacent.
Through observation, we can see that in Equation~\ref{IDM}, all frames are pulled close regardless of time distance.
In Equation~\ref{IDM_2}, all frame features are separated by a margin $m$ if they are not adjacent to the current frame, \ie, all pairs are treated equally.
The above two manners do not fully exploit the smooth and continuous changes of the video.
To this end, we propose a novel form to mine temporal coherence property:
\begin{equation}
\label{temporal_coherence_regularization}
I(\tilde{f_i};\tilde{f_i^a},\tilde{f_i^b}) =
\begin{cases}
  \frac{1}{(a-b)^2+1}\cdot\begin{Vmatrix}\tilde{f_i^a}- \tilde{f_i^b} \end{Vmatrix}, & \text{ if } \left | a-b \right | \le \delta   \\
 max(0, m_{ab}-\begin{Vmatrix}\tilde{f_i^a}- \tilde{f_i^b} \end{Vmatrix}) & \text{ if } \left | a-b \right |>\delta 
\end{cases}
\end{equation}
where $\delta$ is a window size and $ m_{ab}=1-e^{-\frac{(\left | a-b \right |-\delta)^2}{2\sigma^2}}$ for smooth temporal coherence.
Compared with the original forms, our proposed temporal coherence regularization can better reflect the continuous change of video and thus lead to better performance.
%
%
%
%

%
In the training phase, we take the negative distance for each class as logit.
Then we utilize the same cross-entropy loss as in~\citep{OTAM,TRX}, the auxiliary semantic loss~\citep{few-shot-regular,few-shot-regular-2} and the temporal coherence regularization to jointly train the model.
The auxiliary semantic loss refers to the cross-entropy loss on the real action classes, which is widely used to improve training stability and generalization.
During inference, we select the support class closest to the query for classification.
\begin{table*}[t]
\centering
\small
\tablestyle{6pt}{1.0}
\caption{Comparison to recent few-shot action recognition methods on the meta-testing set of SSv2-Full, Kinetics, Epic-kitchens and HMDB51. 
The experiments are conducted under the 5-way setting, and results are reported as the shot increases from 1 to 5.
"-" means the result is not available in published works, and the underline indicates the second best result.
}
\label{tab:compare_SOTA_1}
\begin{tabular}{l|c|c|ccccc}
\shline
Method \hspace{4mm} &  \hspace{1mm} Reference \hspace{1mm} &  \hspace{2mm}  Dataset \hspace{2mm}  &  1-shot   &   2-shot   &   3-shot     &  4-shot     &  5-shot    \\ 
\shline
%
%
CMN++ \cite{CMN}  &  ECCV'18     &  \multirow{14}{*}{SSv2-Full}      & 34.4  & -  & -  & -  & 43.8  \\
TRN++ \cite{TRN-ECCV}  &  ECCV'18     &       & 38.6  & -  & -  & -  & 48.9  \\
OTAM \cite{OTAM}  &   CVPR'20    &        & 42.8  & 49.1  & 51.5  & 52.0  & 52.3  \\ 
TTAN \cite{TTAN}  &   ArXiv'21    &        & 46.3  & 52.5  & 57.3  & 59.3  & 60.4  \\
ITANet \cite{OTAM}  &   IJCAI'21    &        & {49.2}  & \underline{55.5}  &\underline{59.1}  & 61.0  & 62.3  \\ 
TRX ($\Omega{=}\{1\}$) \cite{TRX}    & CVPR'21  &   & 38.8     & 49.7   & 54.4     & 58.0      & 60.6      \\ 
TRX ($\Omega{=}\{2,3\}$)\cite{TRX}  & CVPR'21   &        &  42.0   &  53.1   &  57.6   &   {61.1}   &   {64.6}   \\
STRM \cite{STRM}  & CVPR'22   &        &  43.1   &  53.3   &  \underline{59.1}   &   {61.7}   &   \underline{68.1}   \\
MTFAN~\cite{MTFAN} & CVPR'22 & & 45.7 & - & - & - & 60.4   \\
Nguyen \etal~\cite{nguyen2022inductive} & ECCV'22 & & 43.8 & - & - & - & 61.1   \\
Huang \etal~\cite{huang2022compound} & ECCV'22 & & \underline{49.3} & - & - & - & 66.7   \\
HCL~\cite{zheng2022few} & ECCV'22 & & 47.3 & 54.5 & 59.0 & \underline{62.4} & 64.9   \\

{HyRSM}  & CVPR'22 &   & \hspace{3.7mm} {54.3}\color{light-blue}{ (+5.0)} \hspace{-4mm}  & \hspace{3.7mm} {62.2}\color{light-blue}{ (+6.7)} \hspace{-4mm} & \hspace{3.7mm} {65.1}\color{light-blue}{ (+6.0)} \hspace{-4mm}  &  \hspace{3.7mm} {67.9}\color{light-blue}{ (+5.5)} \hspace{-4mm} & \hspace{3.7mm} {69.0}\color{light-blue}{ (+0.9)} \hspace{-4mm} \\ 
\textbf{HyRSM++}  & - &   & \hspace{3.7mm} \textbf{55.0}\color{blue}{ (+5.7)} \hspace{-4mm}  & \hspace{3.7mm} \textbf{63.5}\color{blue}{ (+8.0)} \hspace{-4mm} & \hspace{3.7mm} \textbf{66.0}\color{blue}{ (+6.9)} \hspace{-4mm}  &  \hspace{3.7mm} \textbf{68.8}\color{blue}{ (+6.4)} \hspace{-4mm} & \hspace{3.7mm} \textbf{69.8}\color{blue}{ (+1.7)} \hspace{-4mm} \\ 
%
\shline
MatchingNet \cite{MatchNet} & NeurIPS'16 &\multirow{18}{*}{Kinetics}      & 53.3  & 64.3  & 69.2  & 71.8  & 74.6  \\ 
MAML \cite{MAML}   & ICML'17   & & 54.2  & 65.5  & 70.0  & 72.1  & 75.3  \\ 
Plain CMN \cite{CMN}  & ECCV'18 & & 57.3  & 67.5  & 72.5  & 74.7  & 76.0  \\ 
CMN-J \cite{CMN-J}  & TPAMI'20 &   &  60.5   &    70.0   &    75.6   &    77.3   &    78.9   \\
TARN \cite{TARN}    & BMVC'19 & & 64.8  & -  & -  & -  & 78.5  \\ 
ARN \cite{ARN-ECCV} &  ECCV'20  &  & 63.7  & -  & -  & -  & 82.4  \\ 
OTAM \cite{OTAM}   & CVPR'20 &  & 73.0  & 75.9  & 78.7  & 81.9  & 85.8  \\ 
ITANet \cite{ITANet}   & IJCAI'21 &  & {73.6}  & -  & -  & -  & 84.3  \\
TRX ($\Omega{=}\{1\}$) \cite{TRX}   & CVPR'21 &  & 63.6  & 75.4  & 80.1  & 82.4  & 85.2  \\ 
TRX ($\Omega{=}\{2,3\}$) \cite{TRX}  & CVPR'21 &   & 63.6  & {76.2}  & {81.8}  & {83.4}  & {85.9}  \\ 
%
%
STRM \cite{STRM}   & CVPR'22 &  & 62.9  & {76.4}  & 81.1  & {83.8}  &  {86.7} \\
MTFAN~\citep{MTFAN} & CVPR'22 & & \textbf{74.6} & - & - & - & \textbf{87.4}   \\
Nguyen \etal~\citep{nguyen2022inductive} & ECCV'22 & & \underline{74.3} & - & - & - & \textbf{87.4}   \\
Huang \etal~\citep{huang2022compound} & ECCV'22 & & 73.3 & - & - & - & 86.4   \\
HCL~\citep{zheng2022few} & ECCV'22 & & 73.7 & \underline{79.1} & \underline{82.4} & \underline{84.0} & 85.8  \\
{HyRSM}  & CVPR'22 &   & \hspace{3.7mm} {73.7}\color{light-red}{ (-0.9)} \hspace{-4mm}  & \hspace{3.7mm} {80.0}\color{light-blue}{ (+0.9)} \hspace{-4mm} & \hspace{3.7mm} \textbf{83.5}\color{light-blue}{ (+1.1)} \hspace{-4mm}  &  \hspace{3.7mm} {84.6}\color{light-blue}{ (+0.6)} \hspace{-4mm} & \hspace{3.7mm} {86.1}\color{light-red}{ (-1.3)} \hspace{-4mm} \\ 
\textbf{HyRSM++}  & - &   & \hspace{3.7mm} \textbf{74.0}\color{red}{ (-0.6)} \hspace{-4mm}  & \hspace{3.7mm} \textbf{80.8}\color{blue}{ (+1.7)} \hspace{-4mm} & \hspace{3.7mm} \textbf{83.9}\color{blue}{ (+1.5)} \hspace{-4mm}  &  \hspace{3.7mm} \textbf{85.3}\color{blue}{ (+1.3)} \hspace{-4mm} & \hspace{3.7mm} {86.4}\color{red}{ (-1.0)} \hspace{-4mm} \\ 
%
\shline 
OTAM \cite{OTAM}  &   CVPR'20    &    \multirow{5}{*}{Epic-kitchens}     & \underline{46.0}  & 50.3  & {53.9}  & 54.9  & 56.3  \\ 
TRX \cite{TRX}  & CVPR'21   &        &  43.4   &  \underline{50.6}   &  53.5   &   {56.8}   &   {58.9}   \\
STRM \cite{STRM}  & CVPR'22   &        &  42.8   &  50.4   &  \underline{54.9}   &   \underline{58.0}   &   \underline{59.2}   \\
{HyRSM}  & CVPR'22 &   & \hspace{3.7mm} {47.4}\color{light-blue}{ (+1.4)} \hspace{-4mm}  & \hspace{3.7mm} {52.9}\color{light-blue}{ (+2.3)} \hspace{-4mm} & \hspace{3.7mm} {56.4}\color{light-blue}{ (+1.5)} \hspace{-4mm}  &  \hspace{3.7mm} {58.8}\color{light-blue}{ (+0.8)} \hspace{-4mm} & \hspace{3.7mm} {59.8}\color{light-blue}{ (+0.6)} \hspace{-4mm} \\ 
\textbf{HyRSM++}  & - &   & \hspace{3.7mm} \textbf{48.0}\color{blue}{ (+2.0)} \hspace{-4mm}  & \hspace{3.7mm} \textbf{54.9}\color{blue}{ (+4.3)} \hspace{-4mm} & \hspace{3.7mm} \textbf{57.5}\color{blue}{ (+2.6)} \hspace{-4mm}  &  \hspace{3.7mm} \textbf{59.6}\color{blue}{ (+1.6)} \hspace{-4mm} & \hspace{3.7mm} \textbf{60.8}\color{blue}{ (+1.6)} \hspace{-4mm} \\ 
%
%
\shline
ARN \cite{ARN-ECCV}  &   ECCV'20    &    \multirow{11}{*}{HMDB51}     & 45.5  & -  & -  & -  & 60.6  \\ 
OTAM \cite{OTAM}  &   CVPR'20    &         & 54.5  & {63.5}  & 65.7  & 67.2  & 68.0  \\
TTAN \cite{TTAN}  &   ArXiv'21    &         & {57.1}  & -  & -  & -  & 74.0  \\  
TRX \cite{TRX}  & CVPR'21   &        &  53.1   &  62.5   &  {66.8}   &   {70.2}   &   {75.6}   \\
STRM \cite{STRM}  & CVPR'22   &        &  52.3   &  62.5   & {67.4}   &   {70.9}   &   \textbf{77.3}   \\
MTFAN~\citep{MTFAN} & CVPR'22 & & 59.0 & - & - & - & 74.6   \\
Nguyen \etal~\citep{nguyen2022inductive} & ECCV'22 & & 59.6 & - & - & - & 76.9   \\
Huang \etal~\citep{huang2022compound} & ECCV'22 & & \underline{60.1} & - & - & - & \underline{77.0}   \\
HCL~\citep{zheng2022few} & ECCV'22 & &59.1 & \underline{66.5} & \underline{71.2} & \underline{73.8} & 76.3  \\
%
{HyRSM}  & CVPR'22 &   & \hspace{3.7mm} {60.3}\color{light-blue}{ (+0.2)} \hspace{-4mm}  & \hspace{3.7mm} {68.2}\color{light-blue}{ (+1.7)} \hspace{-4mm} & \hspace{3.7mm} {71.7}\color{light-blue}{ (+0.5)} \hspace{-4mm}  &  \hspace{3.7mm} {75.3}\color{light-blue}{ (+1.5)} \hspace{-4mm} & \hspace{3.7mm} {76.0}\color{light-red}{ (-1.3)} \hspace{-4mm} \\
\textbf{HyRSM++}  & - &   & \hspace{3.7mm} \textbf{61.5}\color{blue}{ (+1.4)} \hspace{-4mm}  & \hspace{3.7mm} \textbf{69.0}\color{blue}{ (+2.5)} \hspace{-4mm} & \hspace{3.7mm} \textbf{72.7}\color{blue}{ (+1.5)} \hspace{-4mm}  &  \hspace{3.7mm} \textbf{75.4}\color{blue}{ (+1.6)} \hspace{-4mm} & \hspace{3.7mm} {76.4}\color{red}{ (-0.9)} \hspace{-4mm} \\
\shline
\end{tabular}
%
%
\end{table*}

%
\subsection{Extended applications of HyRSM++}
\subsubsection{Semi-supervised few-shot action recognition}
%
%
%
The objective of semi-supervised few-shot action recognition~\citep{CMN-J} is to fully explore the auxiliary information from unlabeled video data to boost the few-shot classification.
Compared with the standard supervised few-shot setting, in addition to the support set $S$ and query set $Q$, an extra unlabeled set $U$ is also included in a semi-supervised few-shot task to alleviate data scarcity.
We demonstrate that the proposed HyRSM++ can build a bridge between labeled and unlabeled examples, leading to higher classification performance.
%

%
Given an unlabeled set $U$, a common practice in semi-supervised learning literature~\citep{zhou2005tri,Mixup,fixmatch} is to adopt the Pseudo Labeling technique~\citep{lee2013pseudo}, which assumes that the decision boundary usually lies in low-density areas and data samples in a high-density area have the same label.
Similarly, traditional semi-supervised few-shot learning methods~\citep{semi-few-shot-1,semi-few-shot-2} usually produce pseudo labels for unlabeled data based on the known support set, and then the generated high-confidence pseudo-label data is augmented into the support set.
In this paper, we follow this paradigm and utilize HyRSM++ to leverage unlabeled examples.
%
Since noisy videos usually have higher losses in training, it is possible to leverage the strong HyRSM++ to distinguish between clean and noisy videos from the prediction scores. 
Based on this, we choose reliable pseudo-labeled samples in the unlabeled set by predictions and augment the support set with high-confidence pseudo-label data.
Subsequently, we take advantage of the augmented support set to classify the query videos as in the supervised few-shot task.
During the training stage, many semi-supervised few-shot tasks are sampled to optimize the whole model, as shown in Algorithm~\ref{semi-algorithm}.
For inference, the evaluation process is also conducted by sampling 10,000 episodic tasks.
%
%
%

\begin{algorithm}[t]
\begin{algorithmic}[1]
\Require{A labeled support set $S$}, an auxiliary unlabeled set $U$, and a query set $Q$
\Ensure{Optimized few-shot classifier HyRSM++  }
\State{Enter support set $S$ and unlabeled set $U$ into HyRSM++ and obtain the category prediction of $U$ based on Equation~\ref{equ:bimhd};}
\State{According to the prediction distribution, select the high-confidence samples to generate pseudo-labels and update $S$ with the selected samples to get the augmented $S^{'}$; }
\State{Apply the augmented $S^{'}$ and query set $Q$ for supervised few-shot training as described in Section~\ref{HyRSM++};}
\end{algorithmic}
\caption{HyRSM++ for semi-supervised few-shot action recognition}
\label{semi-algorithm}
\end{algorithm}

\subsubsection{Unsupervised few-shot action recognition}
%
Unlike the previously described settings involving labelled data, unsupervised few-shot action recognition aims to use unlabeled data to construct few-shot tasks and learn adaptations to different tasks.
We further extend HyRSM++ to this unsupervised task and verify its capability of transferring prior knowledge to learn to deal with unseen tasks efficiently. 
%
%
%

%
To perform unsupervised few-shot learning, constructing few-shot tasks is the first step.
However, there are no label annotations that can be directly applied for few-shot learning in the challenging unsupervised setting.
Following prior unsupervised few-shot algorithms~\citep{UMTRA,ji2019unsupervised}, we generate few-shot tasks by first adopting existing unsupervised learning approaches to learn initialized feature embeddings of the input videos, and then leveraging deep clustering techniques to construct pseudo-classes of the videos.
According to clustering results, we are able to produce few-shot tasks by sampling $N$-way $K$-shot episodes.
We then use the constructed few-shot tasks to train HyRSM++.
During the testing phase, we sample 10,000 episodes from the test set to obtain the performance, and the label information is only used for evaluation.
%
%
%

\section{Experiments}
\label{sec:experiments}
In this section, the following key questions will be answered in detail:
(1) Is HyRSM++ competitive to other state-of-the-art methods on challenging few-shot benchmarks? 
%
(2) What components play an integral role in HyRSM++ so that HyRSM++ can work well?
(3) Can the proposed hybrid relation module be viewed as a simple plug-and-play unit and have the same effect for other methods?
(4) Does the proposed temporal set matching metric have an advantage over other measure competitors?
(5) Can HyRSM++ have stable performance in a variety of different video scenarios?
%

\subsection{Datasets and experimental setups}
\noindent \textbf{Datasets.}
We evaluate our HyRSM++ on six standard public few-shot benchmarks.
For the Kinetics~\citep{Kinetics}, SSv2-Full~\citep{SSV2}, and SSv2-Small~\citep{SSV2} datasets, we adopt the existing splits proposed by~\citep{OTAM,CMN,ITANet,TRX}, and each dataset consists of 64 and 24 classes as the meta-training and meta-testing set, respectively.
For UCF101~\citep{UCF101} and HMDB51~\citep{HMDB51}, we verify our proposed methods by leveraging existing splits from~\citep{ARN-ECCV,TRX}.
In addition to the above, we also utilize the egocentric Epic-kitchens~\citep{EPIC-100-2,EPIC-100} dataset to evaluate HyRSM++. 
%
%

%
%
\begin{table*}[t] 
\centering
\small
\tablestyle{6pt}{1.1}
\caption{Results on 1-shot, 3-shot, and 5-shot few-shot classification on the UCF101 and SSv2-Small datasets. 
%
"-" means the result is not available in published works, and the underline indicates the second best result.
}
\label{tab:compare_SOTA_2}
\begin{tabular}{l|c|ccc|ccc}
\shline
\multicolumn{1}{l}{} & & \multicolumn{3}{c|}{UCF101}  & \multicolumn{3}{c}{SSv2-Small} \\
\shline
\hspace{-0mm} Method \hspace{2mm} &  \hspace{1mm} Reference \hspace{1mm} & \hspace{0mm}  1-shot \hspace{0mm}  &   3-shot & \hspace{3.5mm} 5-shot \hspace{3.5mm}  &  1-shot  & 3-shot & 5-shot \\

\shline
\hspace{-0mm} MatchingNet \cite{MatchNet} & NeurIPS'16       & -  & - & - & 31.3 & 39.8 &  45.5 \\ 
\hspace{-0mm} MAML \cite{MAML}   & ICML'17    & -  & -  & - & 30.9 & 38.6 & 41.9 \\ 
\hspace{-0mm} Plain CMN \cite{CMN}  & ECCV'18  & -  & - & - & 33.4 & 42.5 & 46.5 \\ 
\hspace{-0mm} CMN-J \cite{CMN-J}  & TPAMI'20    &  -   & -  &  - & 36.2 & 44.6 & 48.8  \\
\hspace{-0mm} ARN \cite{ARN-ECCV} &  ECCV'20   & 66.3  & -  & 83.1 & - & - & -  \\ 
\hspace{-0mm} OTAM \cite{OTAM}   & CVPR'20 &  79.9  & 87.0  & 88.9 & 36.4 & 45.9 & 48.0 \\ 
\hspace{-0mm} TTAN \cite{TTAN}   & ArXiv'21 &  {80.9}  & -  & 93.2 & - & - & -  \\
\hspace{-0mm} ITANet \cite{ITANet}   & IJCAI'21   & - & - & - & \underline{39.8} & 49.4 &  53.7 \\
\hspace{-0mm} TRX \cite{TRX}  & CVPR'21 &    78.2  & {92.4}  & {96.1} & 36.0 & \underline{51.9} & \underline{59.1} \\ 
\hspace{-0mm} STRM \cite{STRM}  & CVPR'22 &    80.5  & \underline{92.7}  & \textbf{96.9} & 37.1 & 49.2 & 55.3 \\ %
\hspace{-0mm} MTFAN~\citep{MTFAN} & CVPR'22 & 84.8 & - & 95.1 & - & - &  - \\
\hspace{-0mm} Nguyen \etal~\citep{nguyen2022inductive} & ECCV'22  & \underline{84.9} & - &  \underline{95.9} & - & -  & - \\
\hspace{-0mm} Huang \etal~\citep{huang2022compound} & ECCV'22  & 71.4 & - & 91.0 & 38.9 & - & \textbf{61.6}  \\
\hspace{-0mm} HCL~\citep{zheng2022few} & ECCV'22  &  82.5 & 91.0 & 93.9 & 38.7 &49.1 & 55.4 \\
\hspace{-0mm} {HyRSM}  & CVPR'22 &   \hspace{3.7mm} {83.9}\color{light-red}{ (-1.0)} \hspace{-4mm}  & \hspace{3.7mm} {93.0}\color{light-blue}{ (+0.3)} \hspace{-4mm} & \hspace{3.7mm} {94.7}\color{light-red}{ (-2.2)} \hspace{-3mm} & \hspace{3.7mm} {40.6}\color{light-blue}{ (+0.8)} \hspace{-4mm} & \hspace{3.7mm} {52.3}\color{light-blue}{ (+0.4)} \hspace{-4mm} & \hspace{4mm} {56.1}\color{light-red}{ (-5.5)} \hspace{-3mm} \\ 
\hspace{-0mm} \textbf{HyRSM++}  & - &   \hspace{3.7mm} \textbf{85.8}\color{blue}{ (+0.9)} \hspace{-4mm}  & \hspace{3.7mm} \textbf{93.5}\color{blue}{ (+0.8)} \hspace{-4mm} & \hspace{3.7mm} \underline{95.9}\color{red}{ (-1.0)} \hspace{-3mm} & \hspace{3.7mm} \textbf{42.8}\color{blue}{ (+3.0)} \hspace{-4mm} & \hspace{3.7mm} \textbf{52.4}\color{blue}{ (+0.5)} \hspace{-4mm} & \hspace{4mm} \underline{58.0}\color{red}{ (-2.6)} \hspace{-3mm} \\ 
\shline

\end{tabular}
\end{table*}

\noindent \textbf{Implementation details.}
Following previous works~\citep{CMN,OTAM,TRX,ITANet}, ResNet-50~\citep{Resnet} initialized with ImageNet~\citep{imagenet} pre-trained weights is utilized as the feature extractor in our experiments.
We sparsely and uniformly sample 8 (\ie, $T=8$) frames per video to construct input frame sequence,  which is in line with previous methods~\citep{OTAM, ITANet}.
In the training phase, we also adopt basic data augmentation such as random cropping and color jitter,  
and use Adam~\citep{adam} optimizer to train our model. 
During the inference stage, we conduct few-shot action recognition evaluation on 10,000 randomly sampled episodes from the meta-testing set and report the mean accuracy.
For many shot classification, \eg, 5-shot, we follow ProtoNet~\citep{prototypical} and calculate the mean features of support videos in each class as the prototypes, and classify the query videos according to their distances against the prototypes.
\begin{figure}[t]  
\centering
\includegraphics[width=0.35\textwidth]{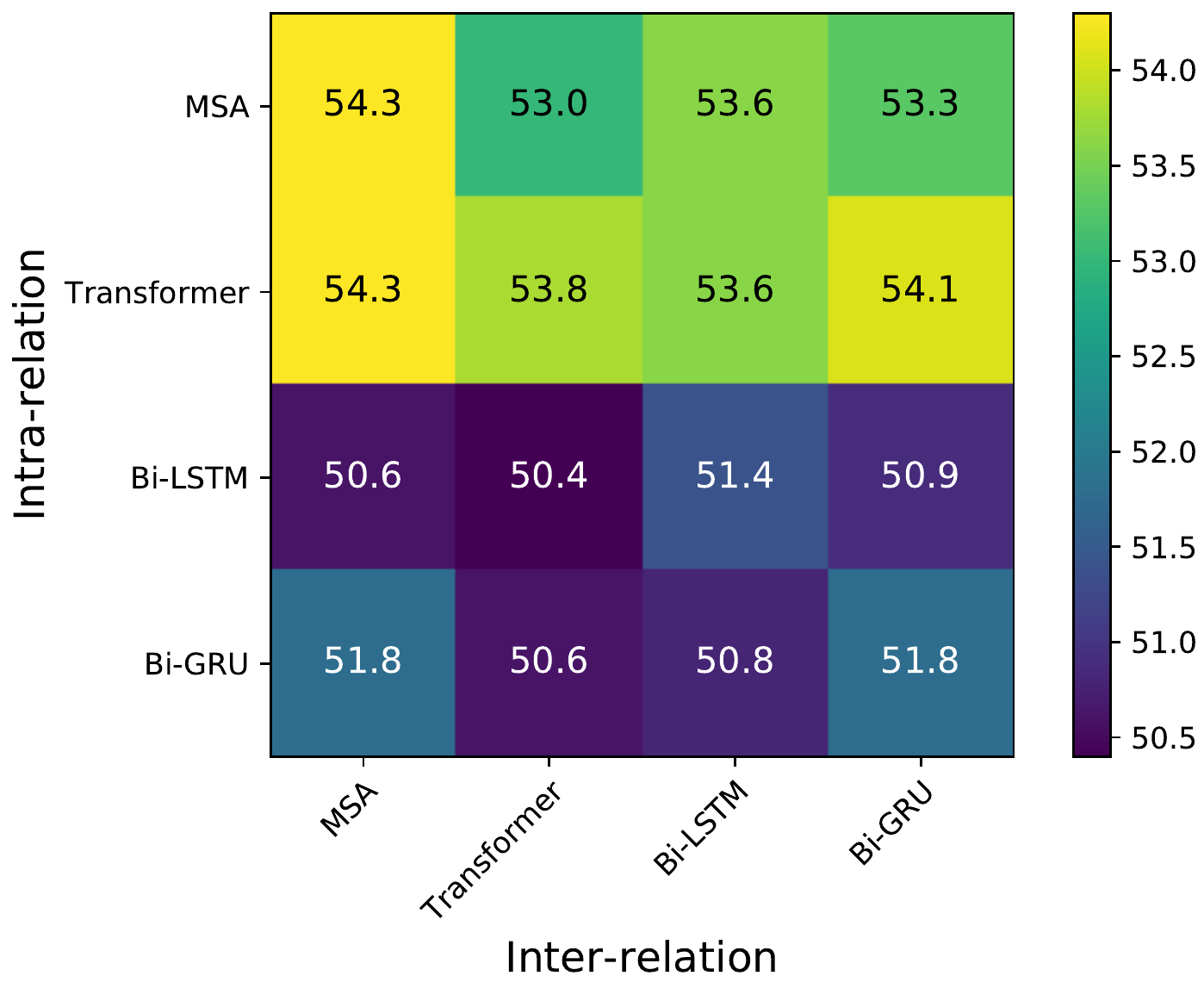}
\caption{Comparison between different components in hybrid relation module on
5-way 1-shot few-shot action classification without temporal coherence regularization. Experiments are conducted on the SSv2-Full dataset.}
\label{fig:intra-intra}
\end{figure}
\begin{figure}[t]  
\centering
\includegraphics[width=0.35\textwidth]{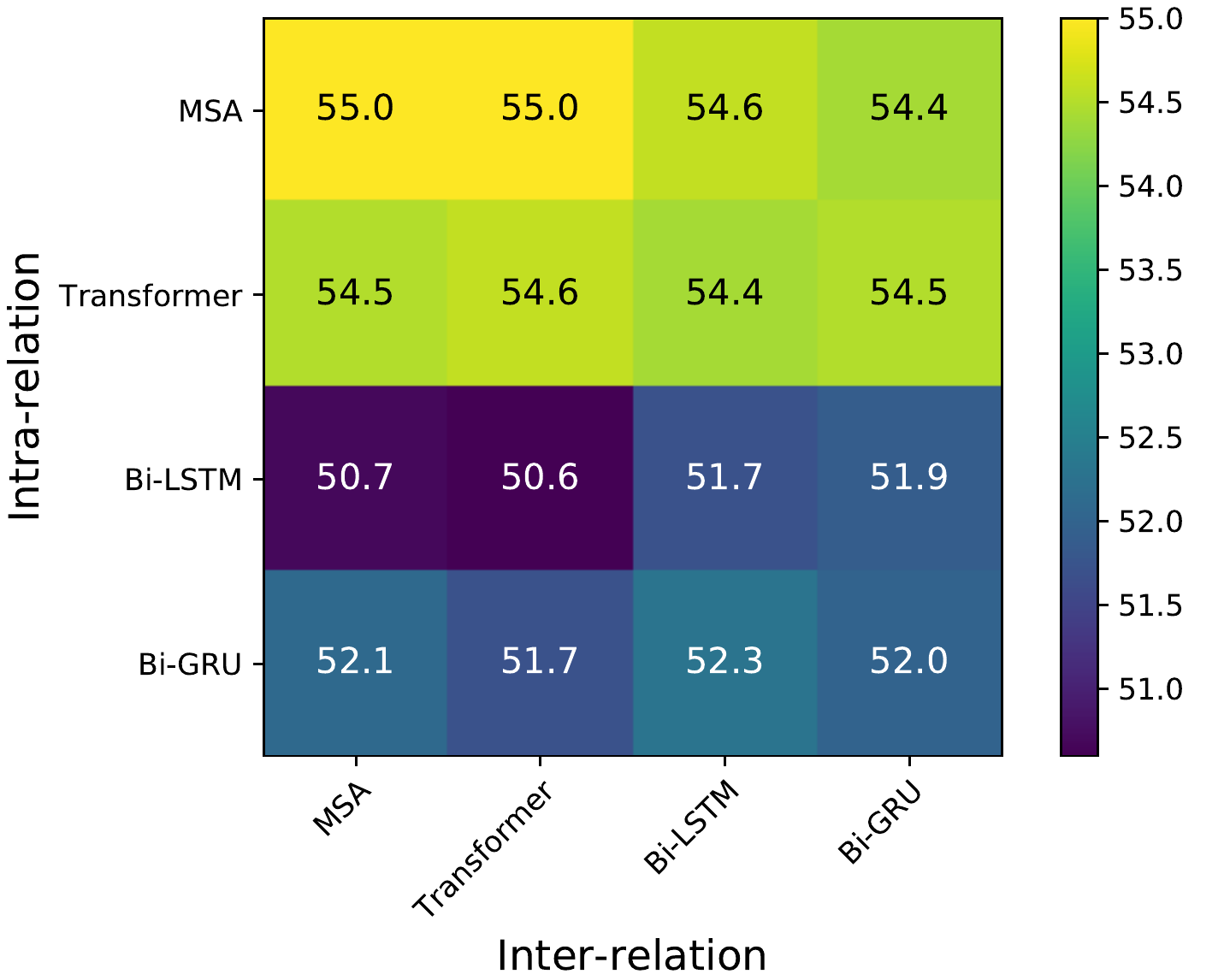}
\caption{Comparison between different components in hybrid relation module on
5-way 1-shot few-shot action classification with temporal coherence regularization. Experiments are conducted on the SSv2-Full dataset.}
\label{fig:intra-intra-plus}
\vspace{-2mm}
\end{figure}
%
%
%
\begin{table}[t]
\centering
\small
\tablestyle{5pt}{1.05}
\caption{Ablation study under 5-way 1-shot and 5-way 5-shot settings on the SSv2-Full dataset. ``TCR" refers to temporal coherence regularization.
}
\label{tab:ablation}
\begin{tabular}{cccc|cc}
\shline
\hspace{-1mm} Intra-relation  &  Inter-relation & Bi-MHM & TCR &   1-shot  &  5-shot \\

\shline
\hspace{-1mm}  &  &  & & 35.2  & 45.3 \\
\shline
\hspace{-1mm} \CheckmarkBold &  &  & & 41.2  & 55.0 \\
\hspace{-1mm}  & \CheckmarkBold &  & & 43.7  & 55.2 \\
\hspace{-1mm}  &  & \CheckmarkBold & & 44.6  & 56.0 \\
\hspace{-1mm}  &  & \CheckmarkBold & \CheckmarkBold & 45.3  & 57.1 \\
\hspace{-1mm} \CheckmarkBold & \CheckmarkBold & &  & 48.1  & 60.5 \\
\hspace{-1mm}  & \CheckmarkBold & \CheckmarkBold & & 48.3  & 61.2 \\
\hspace{-1mm}  & \CheckmarkBold & \CheckmarkBold &\CheckmarkBold & 49.2  & 62.8 \\
\hspace{-1mm} \CheckmarkBold &  & \CheckmarkBold&  & 51.4  & 64.6 \\
\hspace{-1mm} \CheckmarkBold &  & \CheckmarkBold& \CheckmarkBold & 52.4  & 65.8 \\
\hspace{-1mm} \CheckmarkBold & \CheckmarkBold & \CheckmarkBold&  & 54.3  & 69.0 \\
\hspace{-1mm} \CheckmarkBold & \CheckmarkBold & \CheckmarkBold & \CheckmarkBold & \textbf{55.0}  & \textbf{69.8} \\
\shline
\end{tabular}
\end{table}

%
%
%
\begin{table}[t]
\centering
\small
\tablestyle{6pt}{1.1}
\caption{Generalization of hybrid relation module. We conduct experiments on SSv2-Full.
}
\label{tab:generalization}
\begin{tabular}{l|cc}
\shline
\hspace{-1mm} Method  &   1-shot  &  5-shot \\

\shline
\hspace{-1mm}  OTAM~\citep{OTAM}   & 42.8  & 52.3 \\
\hspace{-1mm} OTAM~\citep{OTAM}+ Intra-relation  & 48.9  & 60.4 \\
\hspace{-1mm}  OTAM~\citep{OTAM}+ Inter-relation   & 46.9  & 57.8 \\
\hspace{-1mm} OTAM~\citep{OTAM}+ Intra-relation + Inter-relation & 51.7  & 63.9 \\
\shline
\end{tabular}
\end{table}

\subsection{Comparison with state-of-the-art}
In this section, we validate the effectiveness of the proposed HyRSM++ by comparing it with state-of-the-art methods under various settings. 
%
As indicated in Table~\ref{tab:compare_SOTA_1} and Table~\ref{tab:compare_SOTA_2}, the proposed HyRSM++ surpasses other advanced approaches significantly and is able to achieve new state-of-the-art performance.
For instance, HyRSM++ improves the state-of-the-art performance from 49.2\% to 55.0\% under the 1-shot setting on SSv2-Full and consistently outperforms our original conference version~\citep{HyRSM}.
Specially, extensively compared with current strict temporal alignment techniques~\citep{OTAM,ITANet} and complex fusion methods~\citep{TTAN,TRX}, HyRSM++ produces results that are superior to them under most different shots, which implies that our approach is considerably flexible and efficient.
%
%
%
%
Note that the SSv2-Full and SSv2-Small datasets tend to be motion-based and generally focus on temporal reasoning. 
While Kinetics and UCF101 are partly appearance-related datasets, and scene understanding is usually essential.
%
%
Besides, Epic-kitchens and HMDB51 are relatively complicated and might involve diverse object interactions.
Extensively evaluated on these benchmarks,  HyRSM++ provides excellent performance.
It reveals that our HyRSM++ has strong robustness and generalization for different scenes.
From Table~\ref{tab:compare_SOTA_2}, we observe that HyRSM++ outperforms current state-of-the-art methods on UCF101 and SSv2-Small under the 1-shot and 3-shot settings, which suggests that our HyRSM++ can learn rich and effective representations with extremely limited samples.
It's worth noting that under the 5-shot evaluation, our HyRSM++ yields 95.9\% and 58.0\% 5-shot performance on UCF101 and SSv2-Small, respectively, which is slightly behind STRM and HCL. 
We attribute this to STRM and HCL are ensemble methods that weight each sample with attention or use multiple metrics for few-shot classification, which makes them more suitable for multi-shots, while our HyRSM++ is a simple and general method without involves complex ensemble operations.
Moreover, we also observe that with the introduction of temporal coherence regularization, HyRSM++ has a significant improvement compared to HyRSM, which verifies the effectiveness of exploiting temporal order information during the set matching process.

\subsection{Ablation study}
For ease of comparison, we use a baseline method ProtoNet~\citep{prototypical} that applies global-average pooling to backbone representations to obtain a prototype for each class.
We will explore the role and validity of our proposed modules in detail below.
\begin{figure}[t]  
\centering
\includegraphics[width=0.48\textwidth]{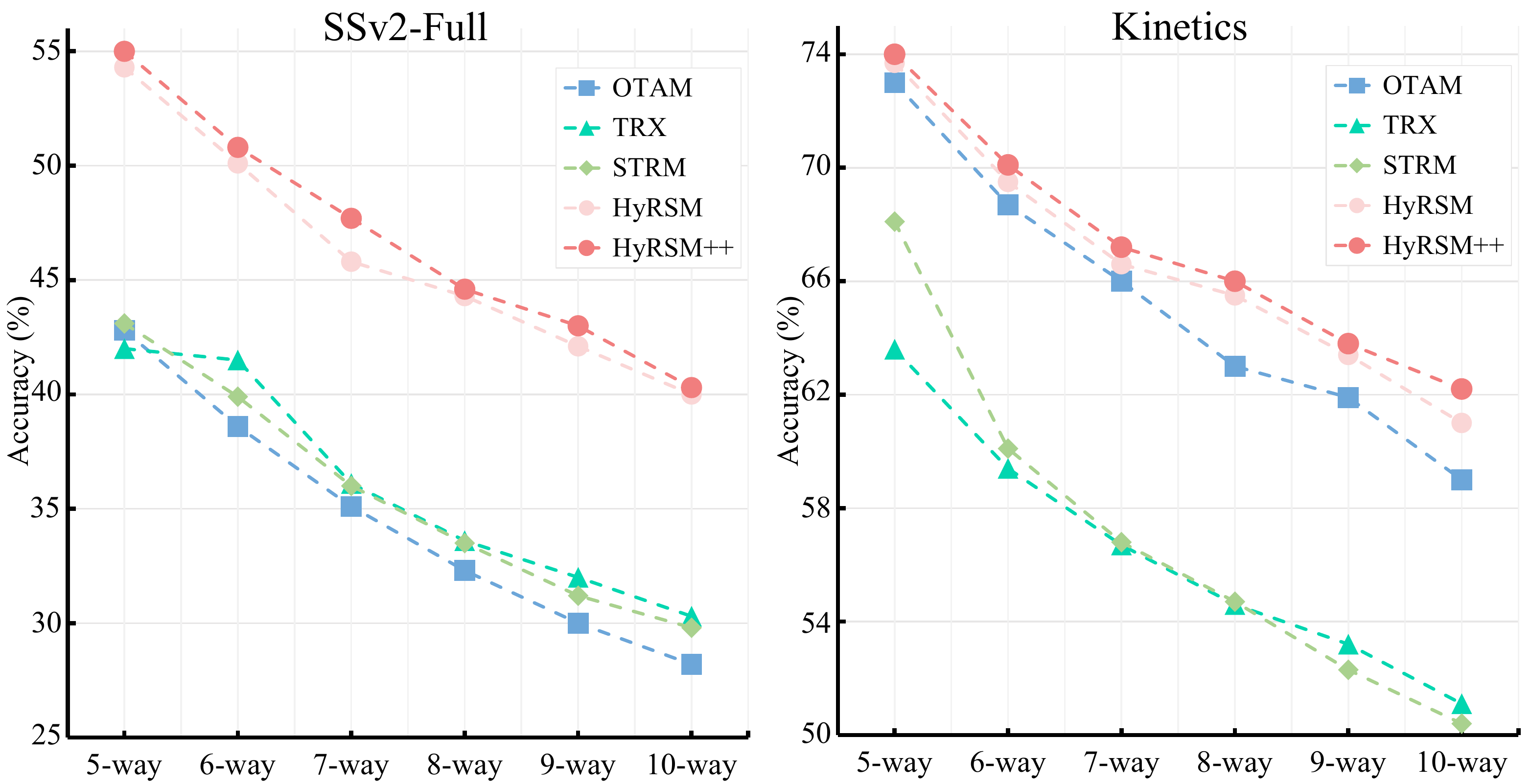}
\caption{N-way 1-shot performance trends of our HyRSM++ and other state-of-the-art methods with different N on SSv2-Full.
The comparison results prove the superiority of our HyRSM++.}
\label{fig:Nway1shot}
\end{figure}

%
%
%
%
%
\begin{figure}[t]  
\centering
\includegraphics[width=0.48\textwidth]{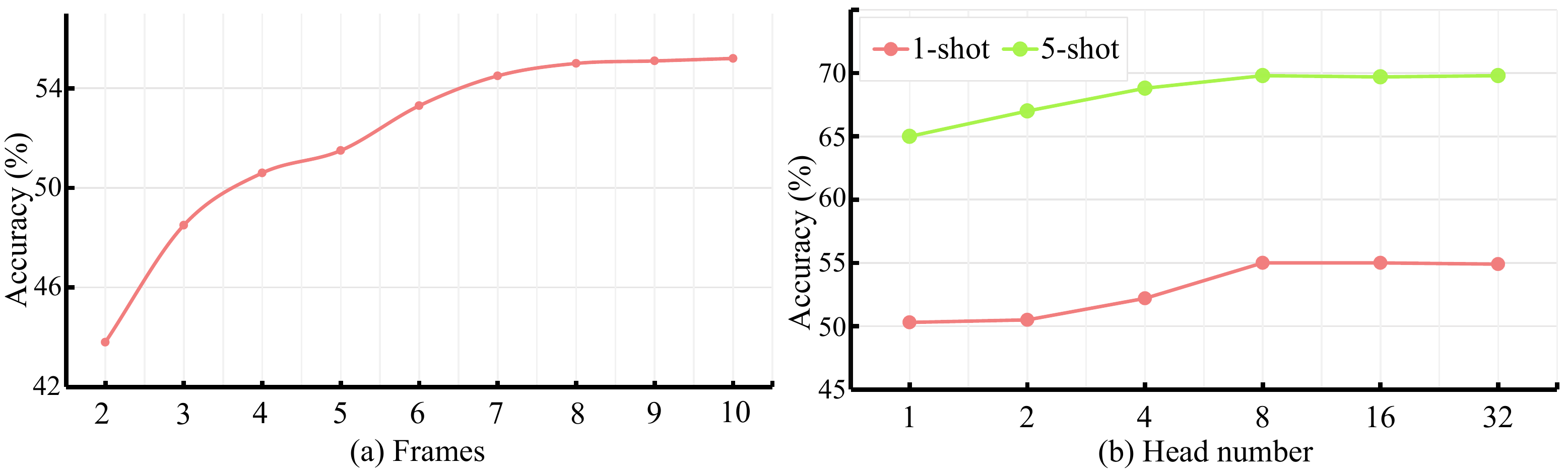}
\caption{(a) Performance on SSv2-Full using a different number of frames under the 5-way 1-shot setting. (b) The effect of the number of heads on SSv2-Full.}
\label{fig:5-way_1-shot_mframes&head}
\end{figure}
%
%
%
\begin{figure}[t]  
\centering
\includegraphics[width=0.48\textwidth]{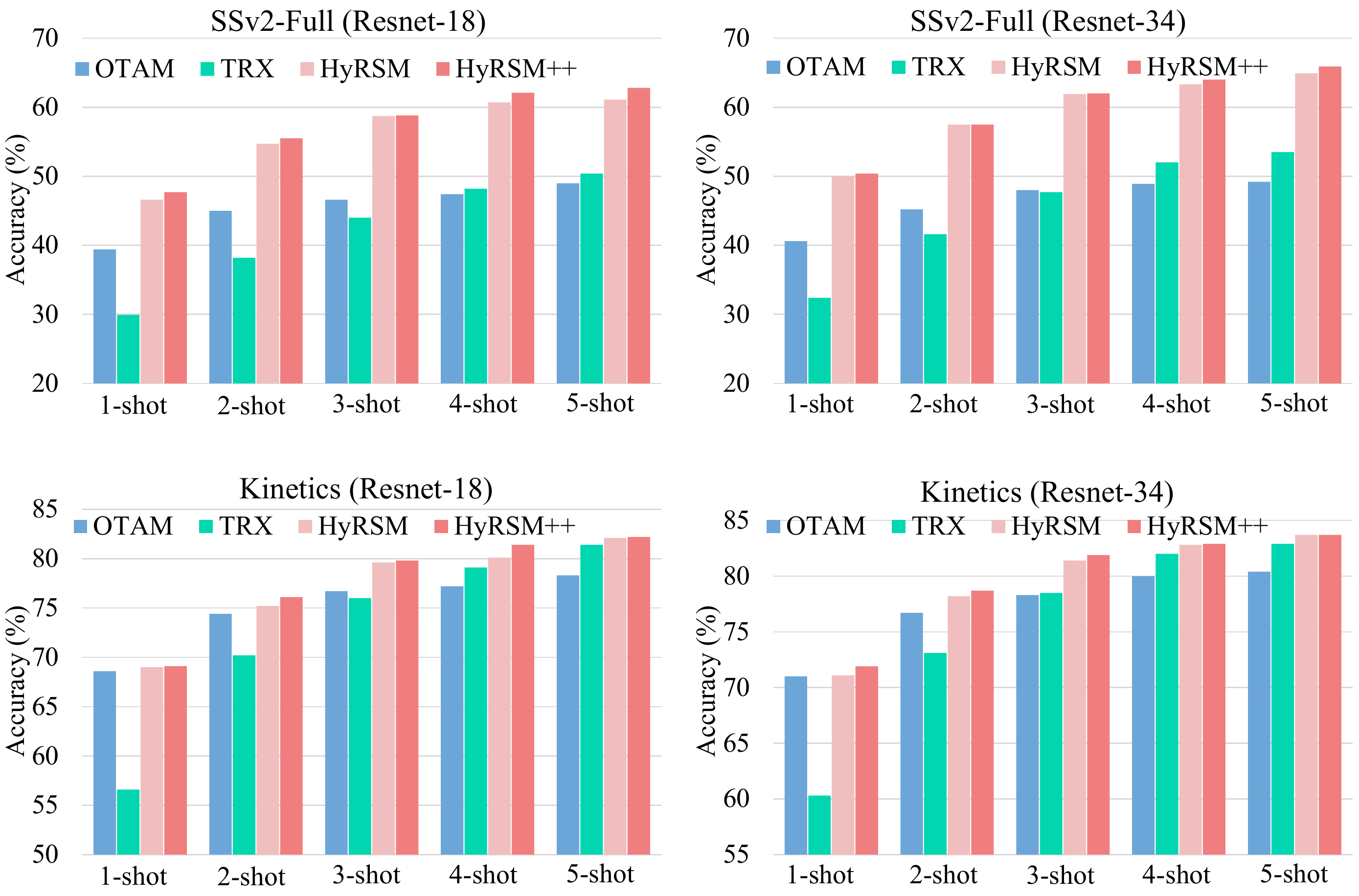}
%
\caption{Comparison of the backbone with different depths on the SSv2-Full and Kinetics datasets.}
\label{fig:resnet18&resnet34}
\end{figure}
\begin{table}[t]
\centering
\small
\tablestyle{6pt}{1.1}
\caption{Comparative experiments on SSv2-Full using the Inception-v3~\citep{Inception-v3} feature extractor.
}
\label{tab:inception-v3}
\begin{tabular}{l|ccccc}
\shline
\hspace{-1mm} Method & 1-shot & 2-shot &  3-shot & 4-shot  &  5-shot \\
\shline
\hspace{-1mm}  OTAM~\citep{OTAM} & 42.4  & 46.6   & 48.7  & 49.2 & 52.1 \\
\hspace{-1mm}  TRX~\citep{TRX} &  37.7  & 50.2 & 55.5  & 57.2 & 60.1 \\
\hspace{-1mm}  STRM~\citep{STRM} &  42.9  & 53.9 & 58.9  & 62.3 & 63.4 \\
\hspace{-1mm}  HyRSM++ & \textbf{53.3} & \textbf{62.7} & \textbf{65.3} & \textbf{67.8}  & \textbf{69.3} \\
\shline

\end{tabular}
\end{table}

\begin{table}[t]
\centering
\small
\tablestyle{6pt}{1.1}
\caption{Performance comparison on SSv2-Full with self-supervised initialization weights~\citep{self-resnet-50}.
}
\label{tab:self-supervised-pretrain}
\begin{tabular}{l|ccccc}
\shline
\hspace{-1mm} Method & 1-shot & 2-shot &  3-shot & 4-shot  &  5-shot \\
\shline
\hspace{-1mm}  OTAM~\citep{OTAM} & 41.2  & 45.9   & 48.8  & 50.1 & 51.0 \\
\hspace{-1mm}  TRX~\citep{TRX} & 37.5  & 43.8 & 49.9  & 51.6 & 52.1 \\
\hspace{-1mm}  STRM~\citep{STRM} & 38.0  & 46.2 & 49.9  & 53.4 & 54.4 \\
\hspace{-1mm}  HyRSM++ & \textbf{50.9} & \textbf{59.1} & \textbf{62.6} & \textbf{65.5}  & \textbf{66.4} \\
\shline

\end{tabular}
\end{table}
%
\begin{table}[t]
\centering
\small
\caption{Performance comparison with different relation modeling paradigms on SSv2-Full and Kinetics.
}
\label{tab:supportset_task}
\tablestyle{6pt}{1.1}
\begin{tabular}{l|c|c|cc}
\shline
\hspace{-1mm} Setting & Method& Dataset &   1-shot  &  5-shot \\

\shline
\hspace{-1mm}  Support-only & HyRSM & \multirow{4}{*}{SSv2-Full}    & 52.1  & 67.2 \\
\hspace{-1mm}  Support-only & HyRSM++ &    & 53.7  & 68.8 \\
\hspace{-1mm}  {Support\&Query}& HyRSM &  & {54.3}  & {69.0} \\
\hspace{-1mm}  \textbf{Support\&Query}& \textbf{HyRSM++} &  & \textbf{55.0}  & \textbf{69.8} \\
\shline
\hspace{-1mm}  Support-only &  HyRSM &\multirow{4}{*}{Kinetics}   & 73.4  & 85.5 \\
\hspace{-1mm}  Support-only &  HyRSM++ &   & 73.5  & 85.7 \\
\hspace{-1mm}  Support\&Query & HyRSM & & {73.7}  & {86.1} \\
\hspace{-1mm}  \textbf{Support\&Query} & \textbf{HyRSM++} & & \textbf{74.0}  & \textbf{86.4} \\
\shline
\end{tabular}
\end{table}
%
%
%
\begin{figure}[t]  
\centering
\includegraphics[width=0.48\textwidth]{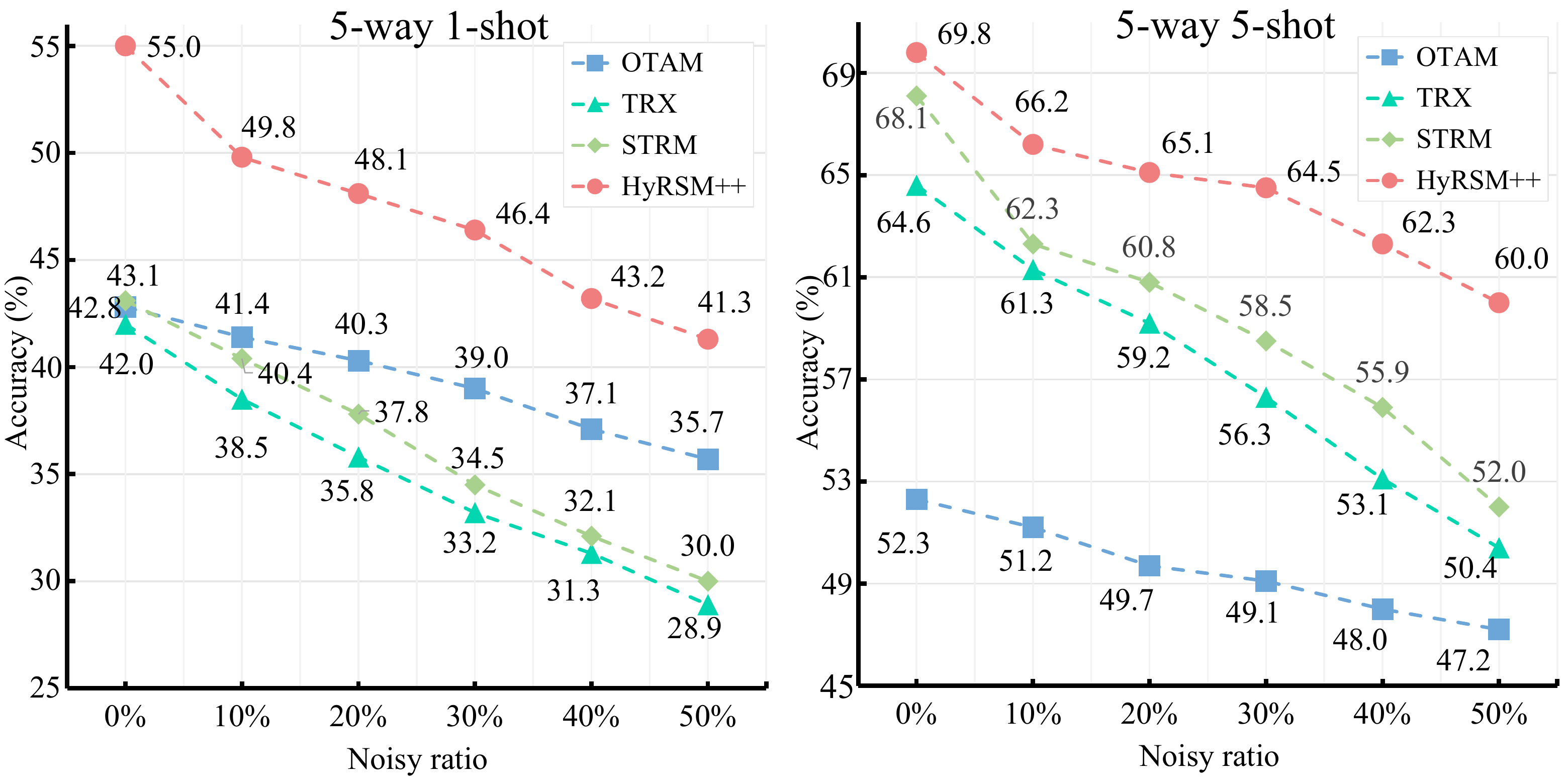}
\caption{Robustness comparison experiments in the presence of noisy samples. X\% represents the proportion of noisy labels included in the dataset.
}
\label{fig:noise}
\end{figure}
%
%
%
\begin{table}[t]
\centering
\small
\tablestyle{6pt}{1.1}
\caption{Comparison with recent temporal alignment methods on the SSv2-Full dataset under the 5-way 1-shot and 5-way 5-shot settings. 
Diagonal means matching frame by frame.
}
\label{tab:matching-alignment}
\begin{tabular}{l|c|cc}
\shline
\hspace{-1mm} Metric \hspace{2mm} &  \hspace{0mm} Bi-direction \hspace{0mm} &   1-shot  &  5-shot \\

\shline
\hspace{-1mm} Diagonal & - &  38.3  & 48.7 \\
\hspace{-1mm} Plain DTW~\citep{DTW} & - &  39.6  & 49.0 \\
\hspace{-1mm} OTAM~\citep{OTAM} & \XSolidBrush &  39.3  & 47.7 \\
\hspace{-1mm} OTAM~\citep{OTAM} & \CheckmarkBold &  {42.8}  & {52.3} \\
\hspace{-1mm} \textbf{Bi-MHM} & \CheckmarkBold & {44.6} & {56.0} \\
\hspace{-1mm} \textbf{Temporal set matching metric} & \CheckmarkBold & \textbf{45.3} & \textbf{57.1} \\
\shline
\end{tabular}
\end{table}

%
\begin{table}[t]
\centering
\small
\tablestyle{6pt}{1.1}
\caption{Comparison of different set matching strategies on the SSv2-Full dataset.
}
\label{tab:matching}
\begin{tabular}{l|c|cc}
\shline
\hspace{-1mm} Metric \hspace{2mm} &  \hspace{0mm} Bi-direction \hspace{0mm} &   1-shot  &  5-shot \\

\shline
\hspace{-1mm} Hausdorff distance & \XSolidBrush & 32.4 & 38.2 \\
\hspace{-1mm} Hausdorff distance & \CheckmarkBold & 34.5 & 39.1 \\
\hspace{-1mm} Modified Hausdorff distance & \XSolidBrush & {44.2} & {50.0} \\
\hspace{-1mm} \textbf{Bi-MHM} & \CheckmarkBold & {44.6} & {56.0} \\
\hspace{-1mm} \textbf{Temporal set matching metric} & \CheckmarkBold & \textbf{45.3} & \textbf{57.1} \\
\shline
\end{tabular}
\end{table}
%
%
%
\begin{table}[t]
\centering
\small
\tablestyle{6pt}{1.1}
\caption{Generalization of temporal coherence regularization. We conduct experiments on SSv2-Full. "Hard margin" represents the method described in Equation~\ref{IDM_2}.
}
\label{tab:temporal_coherence}
\begin{tabular}{l|cc}
\shline
\hspace{-1mm} Method  &   1-shot  &  5-shot \\

\shline
\hspace{-1mm}  OTAM~\citep{OTAM}   & 42.8  & 52.3 \\
\hspace{-1mm}  OTAM~\citep{OTAM} + IDM   & 43.7  & 55.0 \\
\hspace{-1mm}  OTAM~\citep{OTAM} + Hard margin   & 43.2  & 55.3 \\
\hspace{-1mm} OTAM~\citep{OTAM} + Temporal coherence regularization  & {44.1}  & 55.8 \\
\cdashline{1-3}
\hspace{-1mm}  Bi-MHM   & 44.6  & 56.0 \\
\hspace{-1mm}  Bi-MHM + IDM   & 44.7  & 56.3 \\
\hspace{-1mm}  Bi-MHM + Hard margin   & 44.7  & 56.5 \\
\hspace{-1mm} Bi-MHM + Temporal coherence regularization & \textbf{45.3}  & \textbf{57.1} \\
\shline
\end{tabular}
\end{table}
\vspace{+3pt}
\noindent \textbf{Design choices of relation modeling. }
To systematically investigate the effect of different relation modeling operations in hybrid relation module, we vary the components to construct some variants and report the results in  Figure~\ref{fig:intra-intra} and Figure~\ref{fig:intra-intra-plus}.
The comparison experiments are conducted on the  SSv2-Full dataset under the 5-way 1-shot setting.
We can observe that different combinations have quite distinct properties, \eg, multi-head self-attention (MSA) and Transformer are more effective to model intra-class relations than Bi-LSTM and Bi-GRU. 
For example, utilizing multi-head self-attention to learn intra-relation produces at least 2.5\% improvements than with Bi-LSTM.
Nevertheless, compared with other recent algorithms~\citep{TRX,ITANet}, the performance of each combination can still be improved, which strongly suggests the necessity of structure design for learning task-specific features.
%
For simplicity, we choose the same structure to explore intra-relation and inter-relation, and the configuration of multi-head self-attention is adopted in the experiments.

\vspace{+3pt}
\noindent \textbf{Analysis of the proposed components. }
Table~\ref{tab:ablation} summarizes the ablation study of each module in HyRSM++.
To evaluate the function of the proposed components,  ProtoNet~\citep{prototypical} is taken as our baseline.
From the ablation results, we can conclude that each component is highly effective.
In particular, compared to the baseline, intra-relation modeling can respectively bring 6.0\% and 9.7\% performance gains on 1-shot and 5-shot, and inter-relation function boosts the performance by 8.5\% and 9.9\% on 1-shot and 5-shot.
In addition, the proposed set matching metric improves  1-shot and 5-shot classification by 9.4\% and 10.7\%, respectively, which indicates the ability to find better corresponding frames in the video pair.
Adding temporal coherence regularization to the set matching metric also achieves stable performance improvements.
%
Moreover, stacking the proposed modules can further improve performance, indicating the complementarity between components.
When considering all the proposed modules together to form HyRSM++, the performance of 1-shot and 5-shot is improved to 55.0\% and 69.8\%, respectively, which strongly supports the importance of learning task-related features and flexible metrics.


%
%

\vspace{+3pt}
\noindent \textbf{Pluggability of hybrid relation module. }
In Table~\ref{tab:generalization}, we experimentally show that the hybrid relation module generalizes well to other methods by inserting it into the recent OTAM~\citep{OTAM}.
In this study, OTAM with our hybrid relation module benefits from relational information and finally achieves 8.9\% and 11.6\% gains on 1-shot and 5-shot.
This fully evidences that mining the rich information among videos to learn task-specific features is especially valuable.

\vspace{+3pt}
\noindent \textbf{N-way few-shot classification. }
In the previous experiments, all of our comparative evaluation experiments were carried out under the 5-way setting. 
In order to further explore the influence of different N, in Figure~\ref{fig:Nway1shot}, we compare N-way (N $\ge$ 5) 1-shot results on SSv2-Full and Kinetics.
Results show that as N increases, the difficulty becomes higher, and the performance decreases.
Nevertheless, the performance of our HyRSM++ is still consistently ahead of the recent state-of-the-art STRM~\citep{STRM}, TRX~\citep{TRX} and OTAM~\citep{OTAM}, which shows the feasibility of our method to boost performance by introducing rich relations among videos and the power of the set matching metric.

\vspace{+3pt}
\noindent \textbf{Varying the number of frames. }
To demonstrate the scalability of HyRSM++, we also explore the impact of different video frame numbers on performance.
Of note, previous comparisons are performed under 8 frames of input.
Results in Figure~\ref{fig:5-way_1-shot_mframes&head}(a) show that as the number of frames increases, the performance improves. 
HyRSM++ gradually tends to be saturated when more than 7 frames.

\vspace{+3pt}
\noindent \textbf{Influence of head number. }
Previous analyses have shown that multi-head self-attention can focus on different patterns and is critical to capturing diverse features~\citep{multi-head-1}.
We investigate the virtue of varying the number of heads in multi-head self-attention on performance in Figure~\ref{fig:5-way_1-shot_mframes&head}(b). 
Experimental results indicate that the effect of multi-head is remarkable, and the performance starts to saturate beyond a particular point.
%
%
%

%
%
%
%
\begin{figure}[t]  
\centering
\includegraphics[width=0.48\textwidth]{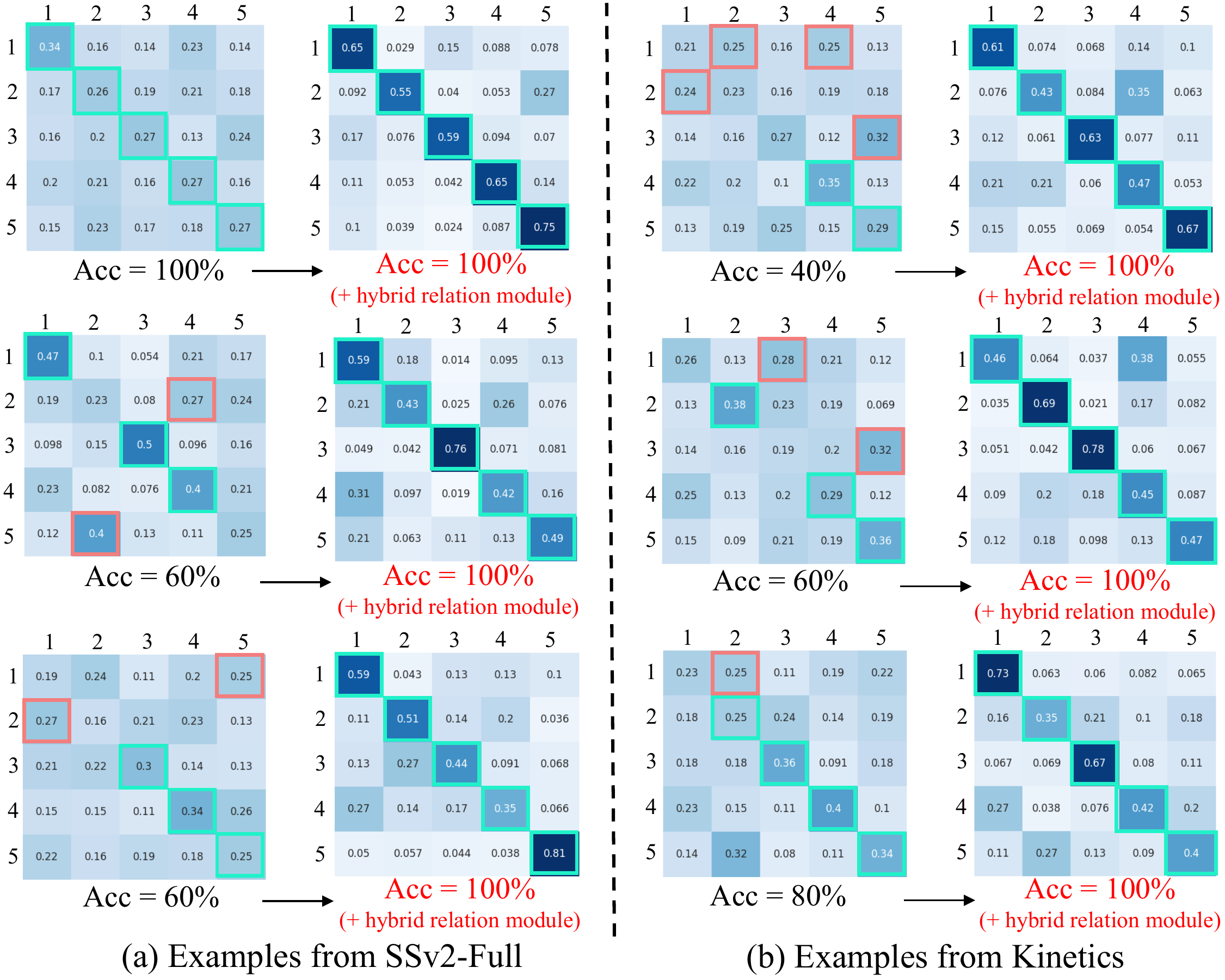}
\caption{Similarity visualization of how query videos (rows) match to support videos (columns). 
The boxes of different colors correspond to: {\color[RGB]{33,241,201}correct match} and {\color[RGB]{242,126,126}incorrect match}.
}
\label{fig:visual_file_relation}
\end{figure}
%
%
%
%
\begin{figure}[t]  
\centering
\includegraphics[width=0.48\textwidth]{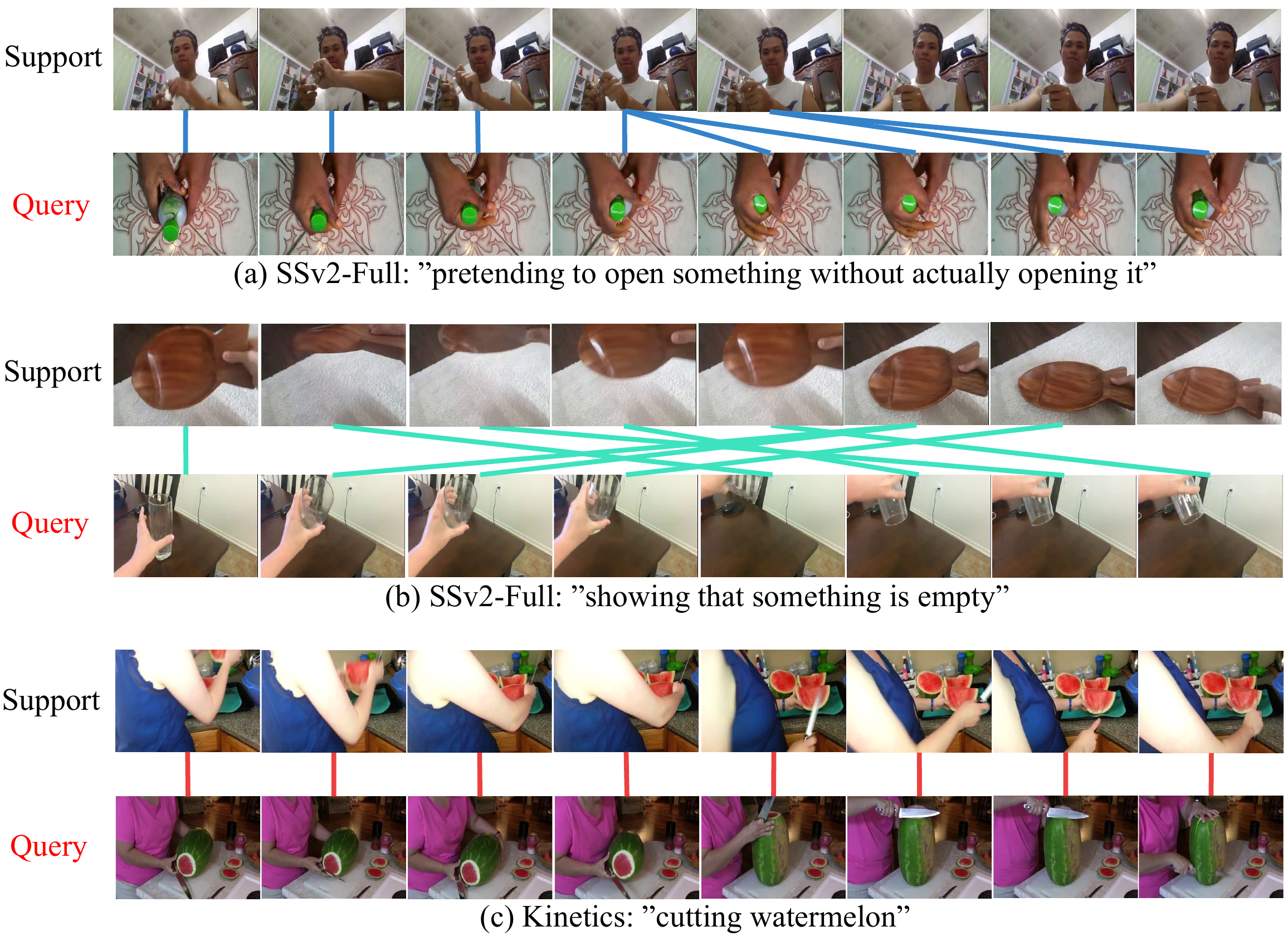}
\caption{Visualization of matching results with the proposed set matching metric on SSv2-Full and Kinetics.
}
\label{fig:set_matching_visual}
\end{figure}

\vspace{+3pt}
\noindent \textbf{Varying depth of the backbone. }
%
%
%
The proposed HyRSM++ is general and compatible with feature extractors of various capacities.
The previous methods all utilize ResNet-50 as backbone by default for a fair comparison, and the impact of backbone's depth on performance is still under-explored.
As presented in Figure~\ref{fig:resnet18&resnet34},  we attempt to answer this question by adopting ResNet-18 and ResNet-34 pre-trained on ImageNet as alternative backbones.
Results demonstrate that the deeper network clearly benefits from greater learning capacity and results in better performance.
In addition, we notice that our proposed HyRSM++ consistently outperforms the competitors (\ie, OTAM and TRX), which indicates that our HyRSM++ is a generally effective framework.

\vspace{+3pt}
\noindent \textbf{Influence of different backbones. }
To verify that our approach is not limited to ResNet-like structures, we further perform experiments on Inception-v3 and report the results in Table~\ref{tab:inception-v3}.
From the comparison, we note that HyRSM++ is significantly superior to other competitive algorithms.
Compared with STRM~\citep{STRM}, our proposed HyRSM++ leads to at least 5.5\% performance gain under various settings.

\vspace{+3pt}
\noindent \textbf{Impact of pretraining types. }
Supervised ImageNet initialization~\citep{imagenet} is widely employed in many vision tasks~\citep{OTAM,CMN-J,OadTR} and achieves impressive success.
Recently, self-supervised techniques have also received widespread attention and revealed excellent application potential.
In Table~\ref{tab:self-supervised-pretrain}, we show the performance comparison with self-supervised pretraining weights~\citep{self-resnet-50}.
Results demonstrate that our HyRSM++ is still powerful and not limited to the specific initialization weights.

\vspace{+3pt}
\noindent \textbf{Other relation modeling forms. }
Previous few-shot image classification methods of learning task-specific features have also achieved promising results~\citep{TapNet,Finding_task-relevant}. %
However, many of them use some complex and fixed operations to learn the dependencies between images, while our method is straightforward and flexible. 
Moreover, most previous works only use the information within the support set to learn task-specific features, ignoring the correlation with query samples.
In our hybrid relation module, we add the query video to the pool of inter-relation modeling to extract relevant information suitable for query classification.
As illustrated in Table~\ref{tab:supportset_task}, we try to remove the query video from the pool in HyRSM++, \ie, \emph{Support-only}, but we can observe that after removing the query video, the performance of 1-shot and 5-shot on SSv2-Full reduces by 1.3\% and 1.0\%, respectively.
There are similar conclusions on the Kinetics dataset.
This evidences that the proposed hybrid relation module is reasonable and can effectively extract task-related features, thereby promoting query classification accuracy.

\vspace{+3pt}
\noindent \textbf{Robustness to noise labels. }
To demonstrate the robustness of HyRSM++ to noise samples, we simulate the presence of noise labels in the dataset in Figure~\ref{fig:noise}.
From the results, we can observe that performance generally decreases as the proportion of noise rises.
However, our HyRSM++ still exhibits higher performance than other methods, which illustrates the robustness of our method and its adaptability to complex conditions.
\begin{table}[bp]
\centering
\small
\tablestyle{6pt}{1.1}
\caption{Complexity analysis for 5-way 1-shot SSv2-Full evaluation. The experiments are carried out on one Nvidia V100 GPU.
}
\label{tab:flops}
\begin{tabular}{lc|cc|cc}
\shline
\hspace{-1mm} Method & Backbone & Param &  FLOPs & Latency  &  Acc \\
\shline
\hspace{-1mm}  HyRSM & ResNet-18 & 13.8M & 3.64G & 36.5ms  & 46.6 \\
\hspace{-1mm}  HyRSM++ & ResNet-18 & 13.8M & 3.64G & 36.5ms  & 47.7 \\
\hspace{-1mm}  HyRSM & ResNet-34 & 23.9M & 7.34G & 67.5ms   & 50.0 \\
\hspace{-1mm}  HyRSM++ & ResNet-34 & 23.9M & 7.34G & 67.5ms   & 50.4 \\
\shline
\hspace{-1mm}  OTAM~\citep{OTAM} & ResNet-50  & \textbf{23.5M}   & \textbf{8.17G}  & 116.6ms & 42.8 \\
\hspace{-1mm}  TRX~\citep{TRX} &  ResNet-50  & 47.1M & 8.22G  & 94.6ms & 42.0 \\
\hspace{-1mm}  STRM~\citep{STRM} &  ResNet-50  & 73.3M & 8.27G  & 113.3ms & 43.1 \\
\hspace{-1mm}  HyRSM & ResNet-50 & 65.6M & 8.36G & \textbf{83.5ms}  & {54.3} \\
\hspace{-1mm}  HyRSM++ & ResNet-50 & 65.6M & 8.36G & \textbf{83.5ms}  & \textbf{55.0} \\
\shline

\end{tabular}
\end{table}
%
%

%
%
%
%
\begin{figure*}[t]  
\centering
\includegraphics[width=0.98\textwidth]{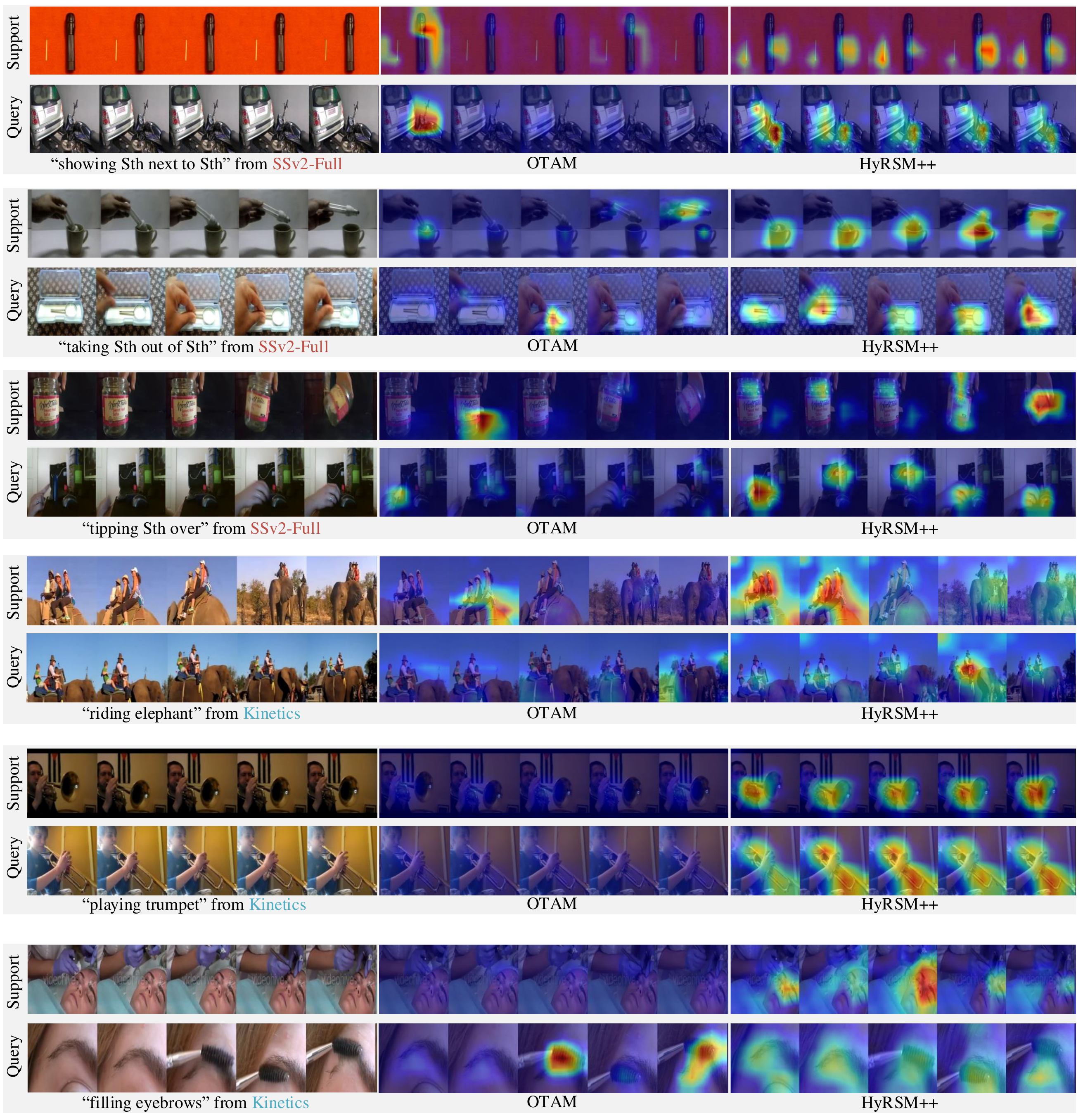}
\caption{Visualization of activation maps with Grad-CAM~\citep{grad_cam}.
Compared to OTAM~\citep{OTAM}, HyRSM++ focuses more precisely on classification-related regions.
}
\label{fig:visual_cam}
\end{figure*}
%
%

\subsection{Comparison with other matching approaches}
%
Our proposed temporal set matching metric Bi-MHM aims to accurately find the corresponding video frames between video pairs by relaxing the strict temporal ordering constraints.
The following comparative experiments in Table~\ref{tab:matching-alignment} are carried out under identical experimental setups, \ie, replace the OTAM directly with our Bi-MHM while keeping other settings unchanged. 
Results show that our Bi-MHM performs well and outperforms other temporal alignment methods (\eg, OTAM).
%
%
We further analyze different set matching approaches in Table~\ref{tab:matching}, and the results indicate that Hausdorff distance is susceptible to noise interference, resulting in the mismatch and relatively poor performance. 
However, our Bi-MHM shows stability to noise and obtains better performance.
Furthermore, compared with the single directional metric,  our proposed bidirectional metric is more comprehensive in reflecting the actual distances between videos and achieves better performance on few-shot tasks.
In addition, we observe that the proposed temporal set matching metric achieves clear improvement over Bi-MHM after incorporating temporal coherence.
For instance, the temporal set matching metric obtains 0.7\%, 1.1\% performance gains on 5-way 1-shot, and 5-way 5-shot SSv2-Full classification. 
It indicates the effectiveness of the proposed temporal set matching metric. 
%
%
%
%
%
\begin{table*}[t]
\centering
\small
\tablestyle{6pt}{1.2}
\caption{Comparison to existing semi-supervised few-shot action recognition methods on the meta-testing set of Kinetics and SSv2-Small. 
The experiments are conducted under the 5-way setting, and results are reported as the shot increases from 1 to 5.
"w/o unlabeled data" indicates that there is no unlabeled set in a episode, \ie, the traditional few-shot action recognition setting, which can act as the lower bound of the semi-supervised counterpart.
}
\label{tab:compare_SOTA_Semi}
\begin{tabular}{l|l|c|ccccc}
\shline
Dataset \hspace{4mm} &   Method \hspace{2mm}  &  \hspace{1mm} Backbone \hspace{1mm}  &  1-shot   &   2-shot   &   3-shot     &  4-shot     &  5-shot    \\ 
\shline
%
%
 \multirow{11}{*}{Kinetics} 
 & OTAM  w/o unlabeled data~\citep{OTAM}  &  Inception-v3    & 68.6  & 72.7 & 74.1  & 75.7  & 76.9  \\
& DeepCluster CACTUs-MAML \cite{CACTUs}  &  Inception-v3     &  65.1     & 72.8  & 76.5  & 77.9  & 79.5    \\
& DeepCluster CACTUs-ProtoNets \cite{CACTUs}  &  Inception-v3    &  66.9     & 73.2  & 77.0  & 78.1  & 79.9    \\
& LIM \cite{CMN-J}  &   Inception-v3         & 69.8  & 75.9  & 78.3  & 80.4  & 82.6  \\
& \textbf{HyRSM++ w/o unlabeled data}  & Inception-v3 & 69.1  & 76.0  & 78.6 & 81.6 &  81.9 \\ 
& \textbf{HyRSM++}  & Inception-v3 & \textbf{73.7}  & \textbf{79.4}  & \textbf{80.9} & \textbf{81.8} &  \textbf{83.1} \\ 
\cdashline{2-8}
& CMN  w/o unlabeled data~\citep{CMN}  &  ResNet-50    & 60.5  & 70.0 & 75.6  & 77.3  & 78.9  \\
& OTAM  w/o unlabeled data~\citep{OTAM}  &  ResNet-50    & 73.0  & 75.9 & 78.7  & 81.9  & 85.8  \\
& LIM (ensemble)~\citep{CMN-J}  &  ResNet-50, Inception-v3, ResNet-18    & 73.3  & 78.3 & 80.8  & 82.4  & 84.0  \\
& \textbf{HyRSM++ w/o unlabeled data}  & ResNet-50 & 74.0  & 80.8  & 83.6 & 85.3 &  86.4 \\ 
& \textbf{HyRSM++}  & ResNet-50 & \textbf{79.1}  & \textbf{84.3}  & \textbf{85.4} & \textbf{86.4} &  \textbf{86.8} \\ 
\shline
%
%
 \multirow{11}{*}{SSv2-Small} 
 & OTAM  w/o unlabeled data~\citep{CMN}  &  Inception-v3    & 36.7  & 41.0 & 43.6  & 44.1  & 46.9  \\
& DeepCluster CACTUs-MAML \cite{CACTUs}  &  Inception-v3     &  37.9     & 44.5  & 45.9  & 47.8  & 49.9   \\
& DeepCluster CACTUs-ProtoNets \cite{CACTUs}  &  Inception-v3    &  38.4     & 44.8  & 46.1  & 48.0  & 50.1    \\
& LIM \cite{CMN-J}  &   Inception-v3         & 41.1  & 46.9  & 48.0  & 51.5  & 53.0  \\
& \textbf{HyRSM++ w/o unlabeled data}  & Inception-v3 & 41.5  & 46.1  & 49.5 & 52.9 &  55.1 \\ 
& \textbf{HyRSM++}  & Inception-v3 & \textbf{43.6}  & \textbf{49.5}  & \textbf{51.8} & \textbf{52.4} &  \textbf{54.5} \\ 
\cdashline{2-8}
& CMN  w/o unlabeled data~\citep{CMN}  &  ResNet-50    & 36.2  & 42.1 & 44.6  & 47.0  & 48.8  \\
& OTAM  w/o unlabeled data~\citep{CMN}  &  ResNet-50    & 36.4  & 42.9 & 45.9  & 46.8  & 48.0  \\
& LIM (ensemble)~\citep{CMN-J}  &  ResNet-50, Inception-v3, ResNet-18    & 44.0  & 49.8 & 51.3  & 53.9  & 55.1  \\
& \textbf{HyRSM++ w/o unlabeled data}  & ResNet-50 & 42.8  & 47.1  & 52.4 & 54.7 &  58.0 \\ 
& \textbf{HyRSM++}  & ResNet-50 & \textbf{45.4}  & \textbf{51.1}  & \textbf{55.2} & \textbf{57.4} &  \textbf{58.8} \\ 
%
%
%
\shline
\end{tabular}
\end{table*}

%
\subsection{Comparison of temporal coherence manners}
%
Pioneering work~\citep{IDM,coherence_unsupervised_1,mobahi2009deep} also indicates the important role of temporal coherence and shows remarkable results in face recognition~\citep{mobahi2009deep} and unsupervised representation learning~\citep{coherence_unsupervised_1,coherence_unsupervised_2}.
However, they also have some limitations as noted in Section~\ref{HyRSM++}, and thus the temporal coherence regularization is proposed for smooth video coherence.
Table~\ref{tab:temporal_coherence} compares the proposed temporal coherence regularization with existing temporal coherence schemes based on OTAM and Bi-MHM.
Results show that exploiting temporal coherence helps improve the classification accuracy of the metrics, which confirms our motivation for considering temporal order information during the matching process.
In addition, our proposed temporal coherence regularization achieves more significant improvements than other manners, and we attribute this to the smooth property of temporal coherence regularization.
%
%

\subsection{Visualization results }
%
%
%
To qualitatively show the discriminative capability of the learned task-specific features in our proposed method, we visualize the similarities between query and support videos with and without the hybrid relation module.
As depicted in Figure~\ref{fig:visual_file_relation}, by adding the hybrid relation module, the discrimination of features is significantly improved, contributing to predicting more accurately.
%
Additionally, the matching results of the set matching metric are visualized in Figure~\ref{fig:set_matching_visual}, and we can observe that our Bi-MHM is considerably flexible in dealing with alignment and misalignment.

To further visually evaluate the proposed HyRSM++, we compare the activation visualization results of HyRSM++ to the competitive OTAM~\citep{OTAM}. 
As shown in Figure~\ref{fig:visual_cam}, the features of OTAM usually contain non-target objects or ignore most  discriminative parts since it lacks the mechanism of learning task-specific embeddings for feature adaptation.
In contrast, our proposed HyRSM++ processes the query and support videos with an adaptive relation modeling operation, which allows it to focus on the different target objects.
The above qualitative experiments illustrate the rationality of our model design and the necessity of learning task-related features.
%

%
%
%
%
\subsection{Limitations}
In order to further understand HyRSM++, Table~\ref{tab:flops} illustrates its differences with OTAM and TRX in terms of parameters, computation, and runtime.
In the inference phase, HyRSM++ does not add additional computational burden compared to HyRSM because the temporal coherence regularization is not involved in the calculation.
Notably, HyRSM++ introduces extra parameters (\ie, hybrid relation module), resulting in increased GPU memory and computational consumption.
Nevertheless, without complex non-parallel classifier heads, the whole inference speed of HyRSM++ is faster than OTAM and TRX.
We will further investigate how to reduce complexity with no loss of performance in the future.

%
%
\begin{figure}[t]  
\centering
\includegraphics[width=0.45\textwidth]{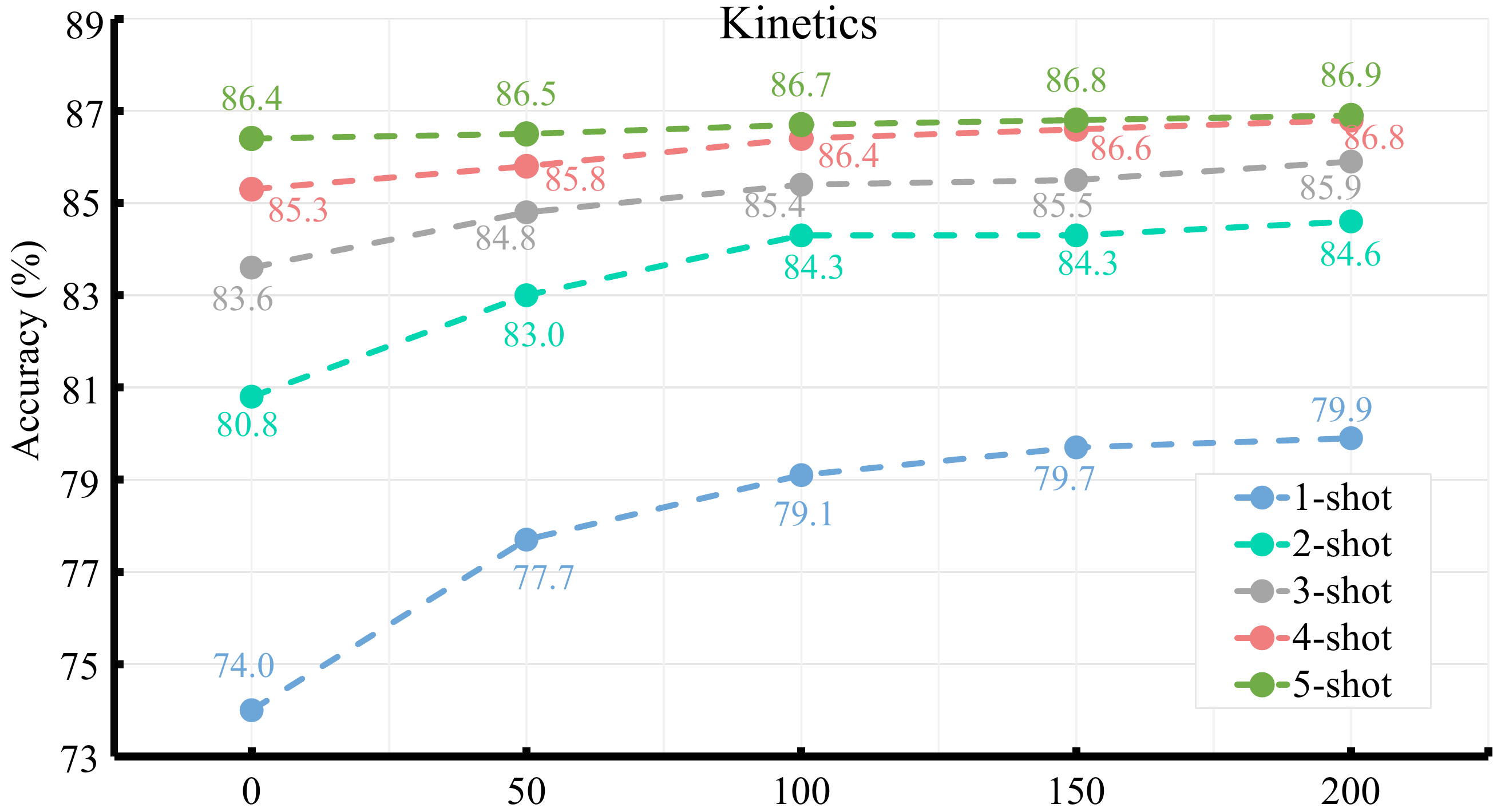}
\caption{Performance comparison of different amounts of unlabeled data for testing in an episode on Kinetics.
%
}
\label{fig:semi_kinetics}
\end{figure}

%
%

%
%
\begin{figure}[t]  
\centering
\includegraphics[width=0.46\textwidth]{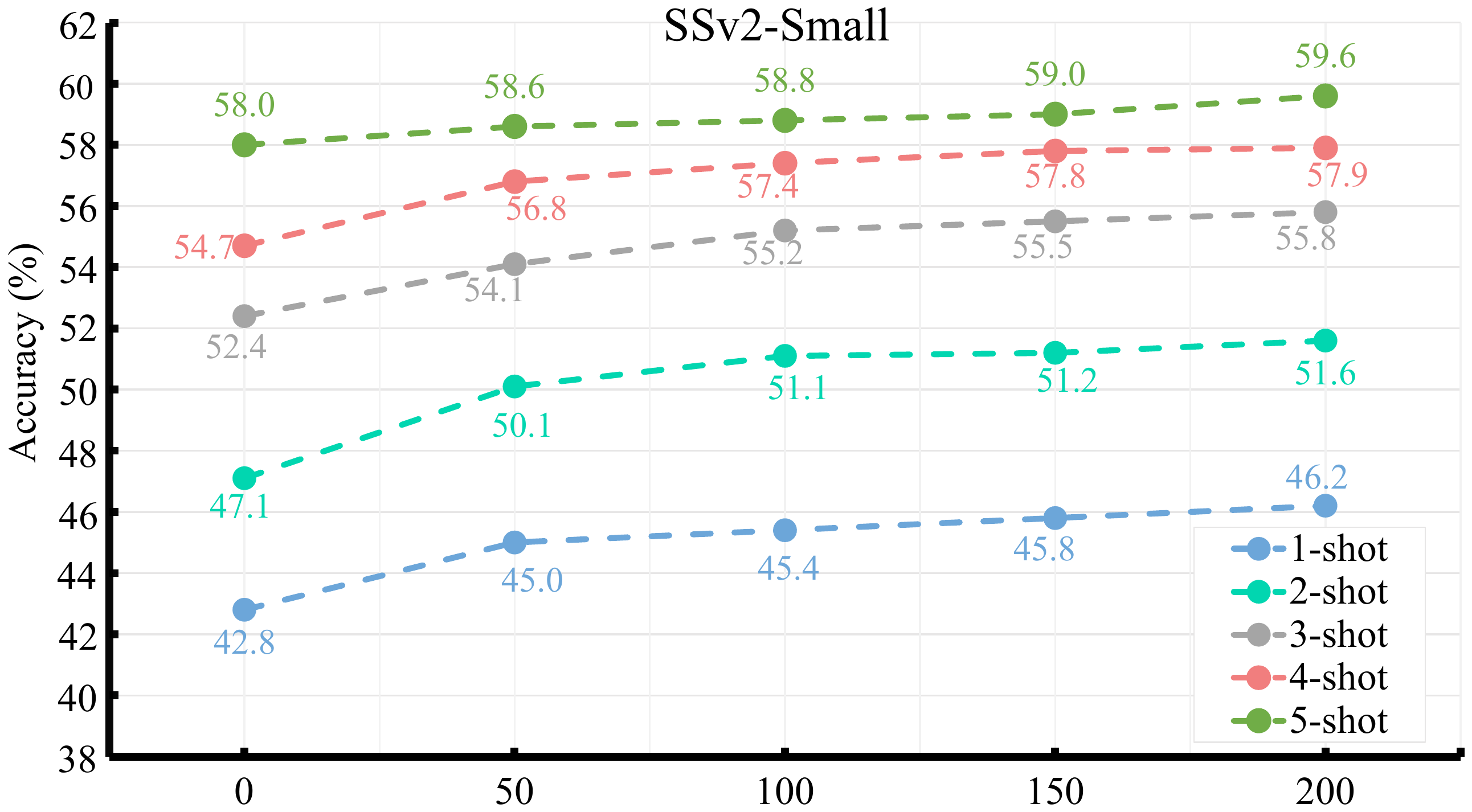}
\caption{Performance comparison of different amounts of unlabeled data for testing in an episode on the SSv2-Small dataset.}
\label{fig:semi_SSv2_small}
\end{figure}

%
%

\begin{table}[t]
\centering
\small
\tablestyle{5pt}{1.1}
\caption{Comparison to state-of-the-art unsupervised few-shot action recognition approaches on UCF101, HMDB51, and Kinetics.
$^{\ast}$ indicates that the algorithm adopt the same 2D ResNet-50 backbone as HyRSM++.
}
\label{tab:compare_with_unsupevised_sota}
\begin{tabular}{l|c|ccc}
\shline
\hspace{-1mm} Method \hspace{-1mm}&\hspace{-1mm} Supervision\hspace{-1mm} & \hspace{-1mm} UCF101 \hspace{-1mm} & \hspace{-1mm} HMDB51\hspace{-1mm}  &\hspace{-1mm} Kinetics \\
\shline
MAML~\citep{MAML} & Supervised & - & - & 54.2   \\
CMN~\citep{CMN} & Supervised & - & - & 60.5   \\
TARN~\citep{TARN} & Supervised & - & - & 66.6   \\
  ProtoGAN~\citep{kumar2019protogan} & Supervised & 57.8 & 34.7 & -   \\
  ARN~\citep{ARN-ECCV} & Supervised & 66.3 & 45.2 & 63.7  \\

\shline
  3DRotNet~\citep{3DRotNet} & Unsupervised  & 39.4   & 32.4  & 27.5  \\
  VCOP~\citep{VCOP} &  Unsupervised  & 32.9 & 27.8  & 26.5  \\
  IIC~\citep{IIC} & Unsupervised & 56.8 & 34.7 & 37.7  \\
  Pace~\citep{Pace_Prediction} & Unsupervised & 25.6 & 26.2 & 22.4  \\
  MemDPC~\citep{IIC} & Unsupervised & 49.3 & 30.3 & 42.0  \\
  CoCLR~\citep{CoCLR} & Unsupervised & 52.0 & 31.3 & 37.6  \\
  {MetaUVFS}$^{\ast}$~\citep{MetaUVFS} & Unsupervised & 66.1 & 40.0 & 50.9  \\
  \textbf{HyRSM++} & Unsupervised & \textbf{68.0} & \textbf{41.0} & \textbf{55.0}  \\
\shline

\end{tabular}
\end{table}

%
%
\begin{figure}[h]  
\centering
\includegraphics[width=0.46\textwidth]{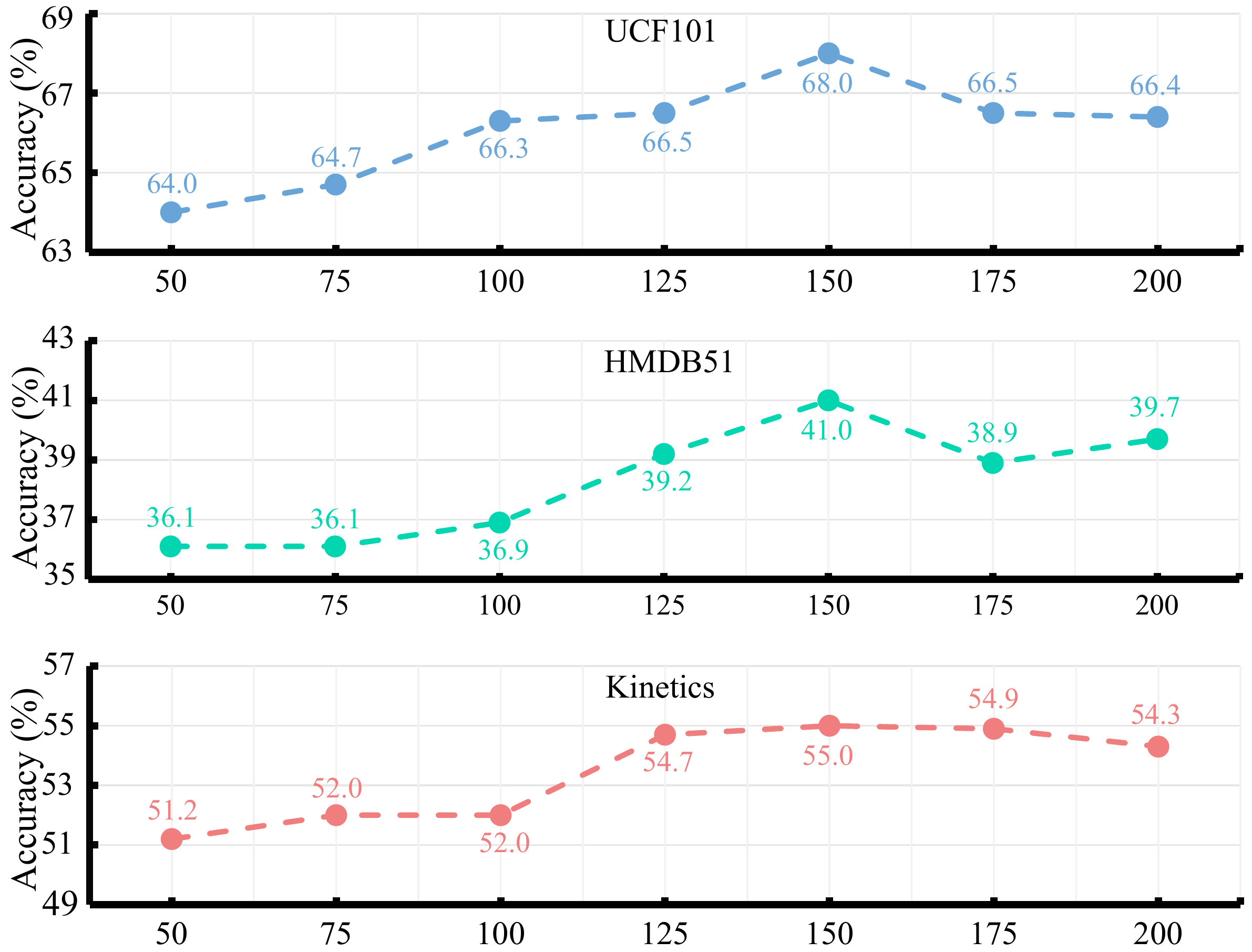}
\caption{Ablation study of different cluster numbers under 5-way 1-shot unsupervised few-shot settings.}
\label{fig:Unsupervised_ablation}
\end{figure}

\section{Extension to Semi-supervised Few-shot Action Recognition}
%
In this section, we demonstrate that the proposed HyRSM++ can be extended to address the more challenging semi-supervised few-shot action recognition problem.
Following LIM~\citep{CMN-J}, we utilize two common datasets (Kinetics~\citep{Kinetics} and SSv2-Small~\citep{SSV2}) to perform comparative experiments.
These two datasets are subsets of Kinetics-400~\citep{Kinetics} and Something-Something-v2~\citep{SSV2}, respectively, and the unlabeled examples in our experiments are collected from the remaining videos of the same category as these subsets.
To conduct the semi-supervised few-shot evaluation, we follow the mainstream distractor setting~\citep{CACTUs,UMTRA,CMN-J}, where the unlabeled set contains other interference classes in each episodic task.
This setting is more realistic and requires the model to be robust to the existence of noisy samples from other classes.
In our experiments, we fixed the number of unlabeled videos in an episodic task to 100.
%
%
%
%
%

Table~\ref{tab:compare_SOTA_Semi} provides the comparison  of our HyRSM++ against state-of-the-art methods on the two standard semi-supervised few-shot benchmarks.
We find that HyRSM++ substantially surpasses the previous approaches, such as LIM~\citep{CMN-J}.
Under the semi-supervised 5-way 1-shot scenario, HyRSM++  produces performance gains of 3.8\% and 2.5\% on Kinetics and SSv2-Small than LIM with Inception-v3 backbone, respectively.
In particular, when using the ResNet-50 backbone, our method is even superior to the multi-modal fusion method (\ie, LIM), which indicates that HyRSM++ enables more accurate pseudo-labels for unlabeled data and then can expand the support set to boost the classification accuracy of the query videos.
In addition, compared to our supervised counterpart (\ie, HyRSM++ w/o unlabeled data), joining unlabeled data is beneficial to alleviating the data scarcity problem and promotes few-shot classification accuracy.
We can observe that when ResNet-50 is adopted as the backbone, the performance of HyRSM++ with unlabeled data is improved by 5.1\% compared to that without unlabeled data under the 5-way 1-shot Kinetics evaluation.
%

%
To further investigate the effect of unlabeled videos in an episode, we conduct comparative experiments with varying numbers of unlabeled videos in Figure~\ref{fig:semi_kinetics} and Figure~\ref{fig:semi_SSv2_small}.
Experimental results show that as the number of unlabeled samples increases, the performance also increases gradually, indicating that the introduction of unlabeled data helps generalize to unseen categories.
%
Furthermore, we notice that the improvement in the 1-shot setting is more significant than that in the 5-shot, which shows that under the condition of low samples, unlabeled videos can improve the estimation of the distribution of new categories more effectively.
Meanwhile, as the amount of unlabeled data increases to a certain level, the performance starts to saturate slowly.
%
%
%

%
\section{Extension to Unsupervised Few-shot Action Recognition}
%
%
We also extend the proposed HyRSM++ to solve the challenging unsupervised few-shot action recognition task where labels for training videos are not available.
Following previous work~\citep{UMTRA,ji2019unsupervised}, we adopt the idea of the "clustering first and then meta-learning" paradigm to construct few-shot tasks and exploit unlabeled data for training.
Our experiments are based on unsupervised ResNet-50 initialization~\citep{self-resnet-50}, which is self-supervised pre-trained on Kinetics-400~\citep{Kinetics} without accessing any label information.
During the clustering process, we utilize the K-means clustering strategy for each dataset to obtain 150 clusters.
%

%
As presented in Table~\ref{tab:compare_with_unsupevised_sota}, we compare HyRSM++ with current state-of-the-art methods on the UCF101, HMDB51 and Kinetics datasets under the 5-way 1-shot setting.
Note that HyRSM++ and MetaUVFS~\citep{MetaUVFS} use the same ResNet-50 structure as the feature extractor, and our HyRSM++ shows better performance on each dataset.
In particular, we observe that our method achieves 68.0\% performance on the UCF101 dataset, a 1.9\% improvement over MetaUVFS, and even surpasses the fully supervised ARN.
The superior performance of HyRSM++ reveals that our approach of leveraging relations within and cross videos and the flexible metric performs effectively in the low-shot regime.
Moreover, this phenomenon also demonstrates the potential of our method to learn a strongly robust few-shot model using only unlabeled videos, even though HyRSM++ is not specifically designed for the unsupervised few-shot action recognition task.
%

%
In the experiments, one parameter involved in applying HyRSM++ to the unsupervised few-shot setting is the number of clusters. 
In Figure~\ref{fig:Unsupervised_ablation}, we display the performance comparison under different number of clusters. 
Results show that when the number of clusters is 150, the performance reaches the peak value, which means that if the cluster number is too small, it may lead to under-clustering. If the number is too large, it may cause over-clustering, damaging the performance.
%
%

\section{Conclusion}
In this work, we have proposed a hybrid relation guided temporal set matching (HyRSM++) approach for few-shot action recognition.
Firstly, we design a hybrid relation module to model the rich semantic relevance within one video and cross different videos in an episodic task to generate task-specific features.
Secondly, built upon the representative task-specific features, an efficient set matching metric is proposed to be resilient to misalignment and match videos accurately.
During the matching process, a temporal coherence regularization is further imposed to exploit temporal order information.
Furthermore, we extend HyRSM++ to solve the more challenging semi-supervised few-shot action recognition and unsupervised few-shot action recognition problems.
%
%
Experimental results demonstrate that our HyRSM++ achieves the state-of-the-art performance on multiple standard benchmarks.

\begin{acknowledgements}
This work is supported by the National Natural Science Foundation of China under grant 61871435, Fundamental Research Funds for the Central Universities no.2019kfyXKJC024, 111 Project on Computational Intelligence and Intelligent Control under Grant B18024, and Alibaba Group through Alibaba Research Intern Program.

\end{acknowledgements}

{\small
\bibliographystyle{spbasic}
\bibliography{egbib}
}
\end{sloppypar}
\end{document}